\documentclass[preprint,12pt]{elsarticle}

\usepackage{amssymb}
\usepackage{booktabs}
\usepackage{epsr}
\usepackage{amsmath}
\usepackage{xcolor}
\usepackage[hyphens]{url}
\usepackage{subcaption}

\usepackage{multirow}

\journal{Electric Power Systems Research}

\begin{document}

\begin{frontmatter}



\title{A multi-stage probabilistic framework to estimate gas-fired generator performance during extreme winter weather}

\author[1]{Sajjad Uddin Mahmud\corref{cor1}}
\ead{sajjaduddin.mahmud@wsu.edu}

\author[1]{Anamika Dubey}
\ead{anamika.dubey@wsu.edu}

\cortext[cor1]{Corresponding author}

\affiliation[1]{organization={School of Electrical Engineering 
               and Computer Science},
               addressline={355 NE Spokane St},
               city={Pullman},
               postcode={99164},
               state={Washington},
               country={USA}}




\begin{abstract}

Extreme winter weather has repeatedly disrupted gas-fired power generation in the United States, yet the plant-level data needed to systematically quantify outage risk remain proprietary. Using publicly available weather and electricity demand data together with anonymized generator contingency records from the North American Electric Reliability Corporation (NERC), we develop a three-stage Bayesian probabilistic framework for estimating winter-driven generator performance. Applied to New York State (2013--2022), the framework sequentially estimates: the hourly probability of a generator contingency event, the expected net available capacity conditioned on an event occurring, and the event duration. Colder conditions and higher electricity demand are associated with higher failure probability, lower retained capacity, and longer event duration. Under the most severe observed stress conditions, estimated mean hourly event probability reaches 24\% , while expected mean net available capacity falls to 13\% of nameplate rating. Full outage events have a median duration of 12.7 hours, while partial derating event duration increases from 2.4 to 7.1 hours with capacity loss severity. The proposed framework establishes a transferable baseline that utilities with access to plant-level records can directly extend to obtain more precise reliability estimates for operational planning and resource adequacy assessment.
\end{abstract}



\begin{highlights}
\item A three-stage Bayesian framework jointly models failure probability, capacity degradation, and event duration of gas-fired generators under extreme winter conditions.
\item A Cold Exposure Index (CEI) constructed from publicly available weather station data captures the persistence and severity of subfreezing conditions at the state level.
\item Under the most extreme winter conditions, hourly failure probability reaches 24\% and — conditional on failure — expected net available capacity of the gas-fired generator falls to 13\% of the nameplate rating.
\item Full outage event duration has a median of 12.7 hours; partial derating event duration increases from 2.4 to 7 hours with capacity loss severity.
\item Despite relying on spatially aggregated, anonymized contingency records, the framework recovers physically consistent relationships between cold exposure, system demand, and generator failure — highlighting the critical need for plant-level data sharing to improve model fidelity for winter  resilience planning and resource adequacy assessment.
\end{highlights}


\end{frontmatter}


\section{Introduction}
Over the past decade, extreme winter weather has become a major source of stress for electric power systems in the United States~\cite{PJM_WinterStormElliott}. For instance, Winter Storm Uri in February 2021 led to 65{,}622\,MW of unplanned generator outages across the South Central United States, left more than four million customers without power, and was linked to more than 200 deaths from cold-related causes~\cite{FERC2021UriReport}. Less than two years later, Winter Storm Elliott in December 2022 caused an even larger coincident generation loss of 90{,}500\,MW across multiple regions~\cite{FERC2023Elliott}. More recently, Winter Storm Fern in January 2026 brought record cold across much of the eastern United States, triggered widespread generator failures, and strained natural gas delivery in the Northeast badly enough that the U.S. Department of Energy issued an emergency order to reduce blackout risks~\cite{DOEFern2026, CRSFern2026}. Across such events, gas-fired generators have repeatedly accounted for a large share of unavailable capacity. Their vulnerability arises from several cold-induced mechanisms, including freezing in pipelines and controls, interruptions in fuel supply, and equipment failures under sustained cold exposure.

Despite the importance of these events, a closer examination of the existing literature reveals limited attention to extreme winter-driven failure analysis~\cite{kasimalla2024evaluation, mishra2024review}. Recent work on power system resilience and planning under extreme weather has largely focused on developing component-level impact models, operational and planning strategies under hazards such as extreme winds, hurricanes, floods and wildfires~\cite{sturmer2024increasing, mujjuni2023evaluation, abodhjournal, panteli2015modeling, wang2023resilience, chowdhury2026resilient, nikoobakht2022collaborative, khan2025aggregated}. Even in studies that focus on weather-driven impacts across power system components, extreme cold and snow conditions are mainly associated with load variations, reduced renewable output, and transmission or distribution system impacts~\cite{xu2024resilience, gutierrez2026power, cerrai2020outage, lv2026risk}. The response of thermal generation under sustained cold exposure remains comparatively underexplored. Existing studies that consider generator outages typically analyze performance across broad weather and system loading conditions, without explicitly isolating extreme winter effects~\cite{murphy2018resource, murphy2019time}. More broadly, data-driven frameworks linking extreme weather to power system failure have advanced considerably for heatwaves, compound warm-season events, and 
wind-driven hazards~\cite{saki2025heatwave, sturmer2024increasing}, yet no equivalent study exists for winter weather-driven generator contingencies. Attempts at winter storm scenario generation rely on fitting outage distributions to a single historical event rather than a predictive mapping to cold exposure and system demand   ~\cite{austgen2022winter}. Physics-based models can capture cold-weather effects at the component level~\cite{ge_gas_turbine_performance}, but they require detailed plant-specific information and are difficult to scale to system-level analysis across large generator fleets. As a result, a scalable, data-driven framework that directly connects cold exposure and system demand to quantify generator failure risk under extreme winter conditions is still lacking.

We address this gap using the Generating Availability Data System (GADS), maintained by the North American Electric Reliability Corporation (NERC). GADS records operational interruption events and performance for generating units of 20 MW and above across the United States and Canada since 2013~\cite{GADS}. Cold exposure is a primary driver of winter-related generator failures, as many gas-fired units are affected by freezing of equipment and fuel supply systems. At the same time, electricity demand typically increases during extreme cold events, placing additional stress on the system. We therefore focus on these two variables as the key drivers in the model. The version of GADS data available for research use does not contain power plant identifiers. Generator locations are anonymized to the state level, so temperature and electricity demand can only be paired with generator contingencies at the state level. To mitigate this limitation, we focus on New York as a representative test case in this study. The main contributions of this paper are as follows:

\begin{itemize}
\item A cold exposure index (CEI) is constructed from publicly available weather station data to represent the persistence and severity of subfreezing conditions experienced by gas-fired generators at the state level.

\item A three-stage Bayesian probabilistic framework is developed to sequentially estimate hourly event probability, conditional net available capacity as functions of CEI and electricity demand and event duration as a function of event severity. A Bayesian formulation is well suited to this problem because extreme winter-driven events are rare and the available data are spatially aggregated, making 
it important to explicitly account for uncertainty rather than rely on point 
estimates~\cite{sajjadpesgm25}.

\item The framework is applied on New York State GADS records, demonstrating that colder conditions and higher demand are jointly associated with elevated failure risk, lower retained capacity, and longer event duration.

\item The proposed approach is designed to be transferable: utilities with access to plant-level records can directly extend the framework to obtain higher-resolution reliability estimates for operational planning and resource adequacy assessment.

\end{itemize}

The remainder of this paper is organized as follows. Section~2 describes the data preparation and the proposed three-stage Bayesian framework. Section~3 presents the test case and model parameters. Section~4 shows the results for New York State. Section~5 discusses the implications and limitations of the framework with concluding remarks.

\section{Data Preparation and Proposed Framework}\label{sec2}

\subsection{Generator contingency data and winter driven event identification}\label{sec2_gadsdata}
NERC GADS data contains generator contingency event data that includes start and end timestamps, event type, cause code, amplification code, fuel type, and the normalized net available capacity ($\mathrm{NAC_{norm}}$) during the event~\cite{gadsreporting}. $\mathrm{NAC_{norm}}$ is defined as the ratio of net available capacity to net maximum capacity. Based on $\mathrm{NAC_{norm}}$, events are classified as full outages ($\mathrm{NAC_{norm}} = 0$) or partial derating events ($0 < \mathrm{NAC_{norm}} < 1$). To extract winter-weather-driven gas-fired contingency events, we apply a structured filtering procedure. We first consider unplanned/forced contingency events for generator with primary fuel as gas. Unplanned event types include immediate, delayed, and postponed forced outages (U1, U2, U3), the corresponding forced deratings (D1, D2, D3), and startup failures (SF). We then restrict to events occurring during winter months (December--February) within the study period. To identify weather-driven events, we retain records carrying cause code~9036 (``Storms: ice, snow, etc.'') or amplification code~48 (``Frozen: temperature-related''). Amplification code~48 is essential because many cold-driven equipment related failures are recorded under equipment-level cause codes with no explicit weather label. Relying on cause code~9036 alone would substantially undercount winter-driven contingencies. Even with both codes combined, the filtering remains conservative because amplification codes are optional in GADS reporting and are not consistently reported by plant operators.

\subsection{Weather--generator mapping and cold-exposure index}\label{sec2_weatherdata}

A key limitation of the available GADS data is that generator locations are anonymized to the state level and plant identifiers are withheld. As a result, temperature can only be paired with generator contingencies at the state level. Hence, we construct an aggregated state-level temperature profile to approximate cold exposure of the generators using observations from the Integrated Surface Database (ISD) maintained by the National Oceanic and Atmospheric Administration (NOAA)~\cite{NOAA_ISD}. ISD includes weather stations that may be geographically distant from any gas-fired generators. Therefore, we restrict the aggregation to stations located near gas-fired power plants by using gas-fired power plant locations from the Emissions \& Generation Resource Integrated Database (eGRID)~\cite{egrid}. This procedure removes geographically irrelevant stations and retains a subset of weather stations that are representative of generator locations. 

For each gas-fired power plant $p$ in a state $s$, the nearest station $w(p)$ is identified by
\begin{equation}
w(p) = \arg\min_{w \in W_s} d(p,w)
\end{equation}
where $W_s$ is the set of ISD stations in state $s$, and $d(p,w)$ is the great-circle distance between plant $p$ and station $w$. The aggregation set is
\begin{equation}
W_s^{\ast} = \bigcup_{p \in P_s} w(p)
\end{equation}
where $P_s$ is the set of gas-fired power plants in state $s$ in eGRID. Although eGRID and GADS describe the same underlying fleet, plant-level identifiers are anonymized in the available GADS data, preventing direct linkage. Accordingly, eGRID is used only to select representative nearby weather stations, while all subsequent modeling is conducted at the state level, consistent with the resolution of GADS. Then the hourly state-level aggregated temperature is defined as the minimum across the selected weather stations:
\begin{equation}
\overline{T}^{\min}_{s,h} = \min_{w \in W_s^{\ast}} T_{w,h}
\label{eq:temp_agg1}
\end{equation}
where $T_{w,h}$ is the recorded temperature at station $w$ during hour $h$ in state $s$. We use the minimum to provide a conservative representation of cold-weather stress.

Because winter-driven generator failures depend not only on instantaneous temperature but also on the persistence of sub-freezing conditions, we define a cold exposure index (CEI) in this study. For each hour $t$ in the dataset, a 25-hour temperature window is assembled from the current-hour temperature $T_t$ and the preceding 24-hour profile $\{T_{t-1}, T_{t-2}, \ldots, T_{t-24}\}$.
The raw CEI accumulates sub-freezing exposure over this window using
exponentially decaying weights, so that more recent cold contributes more
than cold that occurred a day prior:
\begin{equation}
\mathrm{CEI}_{raw}
= \max(0,\,-\overline{T}^{\min}_{s,t})
+ \sum_{k=1}^{24} \lambda^k\,\max(0,\,-\overline{T}^{\min}_{s,t-k})
\label{eq:cei_raw}
\end{equation}
where $T$ is in $^\circ$C (so the freezing temperature is $0\,^\circ$C), and
$\lambda \in (0,1)$ is the recency decay parameter.
The term $\max(0,-T)$ equals the number of degrees below freezing at a given
hour, and is zero for any above-freezing hour. Prior to model fitting, the $\mathrm{CEI}_{raw}$ is standardized using the mean $\mu^{\mathrm{train}}_{\mathrm{CEI}}$ and standard deviation $\sigma^{\mathrm{train}}_{\mathrm{CEI}}$ that are computed from the training data:
\begin{equation}
{\mathrm{CEI}}
= \frac{\mathrm{CEI}_{raw} - \mu^{\mathrm{train}}_{\mathrm{CEI}}}
       {\sigma^{\mathrm{train}}_{\mathrm{CEI}}}
\label{eq:cei_scaled}
\end{equation}

\subsection{Demand data aggregation}\label{sec2_demanddata}
Electricity demand is used as the second covariate because system loading amplifies generator stress during extreme cold events. As generator-level loading data are not available in GADS, publicly available zonal-level hourly electricity demand is obtained from the corresponding Independent System Operator (ISO). This hourly demand is then normalized within each winter season before being used as a model input. The winter season of year $Y$, denoted $\mathcal{W}_Y$, comprises December of year $Y\!-\!1$ and January--February of year $Y$. Normalized demand is defined as: 
\begin{equation}
D_{\mathrm{norm}} = \frac{D_t^Y}{D^Y_{\text{peak}}},
\qquad t \in \mathcal{W}_Y
\end{equation}
where $D_t^Y$ is the hourly demand and $D^Y_{\mathrm{peak}}$ is the peak demand of the winter season $Y$, giving $D_{\mathrm{norm}} \in (0,1]$. $D_{\mathrm{norm}}$ represents the aggregate loading conditions experienced by the generator fleet in the model. 

\subsection{Hourly dataset for probabilistic modeling}\label{sec2_hourlydata}

The raw GADS dataset contains only event-level records of generator full outage and partial derating and does not include hours without events. To enable probabilistic modeling, we convert these records into an hourly time series and merge them with $\mathrm{CEI}$ and $D_\mathrm{norm}$ data. For each winter season, we construct a continuous hourly timeline covering December, January, and February. Let $\mathcal{H}$ denote the set of all hourly timestamps in a given season. This timeline includes all hours, regardless of whether a contingency event occurs. Each GADS event is defined by a start time and an end time. To identify the onset of events, we define a binary indicator $E_t$ for each hour $t \in \mathcal{H}$ as
\begin{equation}
E_t =
\begin{cases}
1, & \text{if a new contingency event is initiated during hour } t \\
0, & \text{otherwise}
\end{cases}
\end{equation}
An event is considered to be initiated in hour $t$ if its recorded start time falls within that hour. Hours during which events are ongoing but not newly initiated are assigned $E_t = 0$, as the objective of Stage~1 is to model event initiation. For each hour $t$, the binary event indicator $E_t$ is merged with the corresponding ($\mathrm{CEI}$) and $D_{\mathrm{norm}}$ variables.  This results in a structured hourly dataset of the form
\[
\left\{ (E_t, \mathrm{CEI}_t, D_{\mathrm{norm},t}) \;:\; t \in \mathcal{H} \right\},
\]
which includes both event and non-event observations.  This hourly dataset is used to estimate the probability of event initiation (Stage~1) as a function of $\mathrm{CEI}$ and $D_\mathrm{norm}$.

\subsection{Proposed Three-Stage Bayesian Framework}\label{sec2_framework}
\begin{figure*}[t!]
\centering
\includegraphics[width=\textwidth]{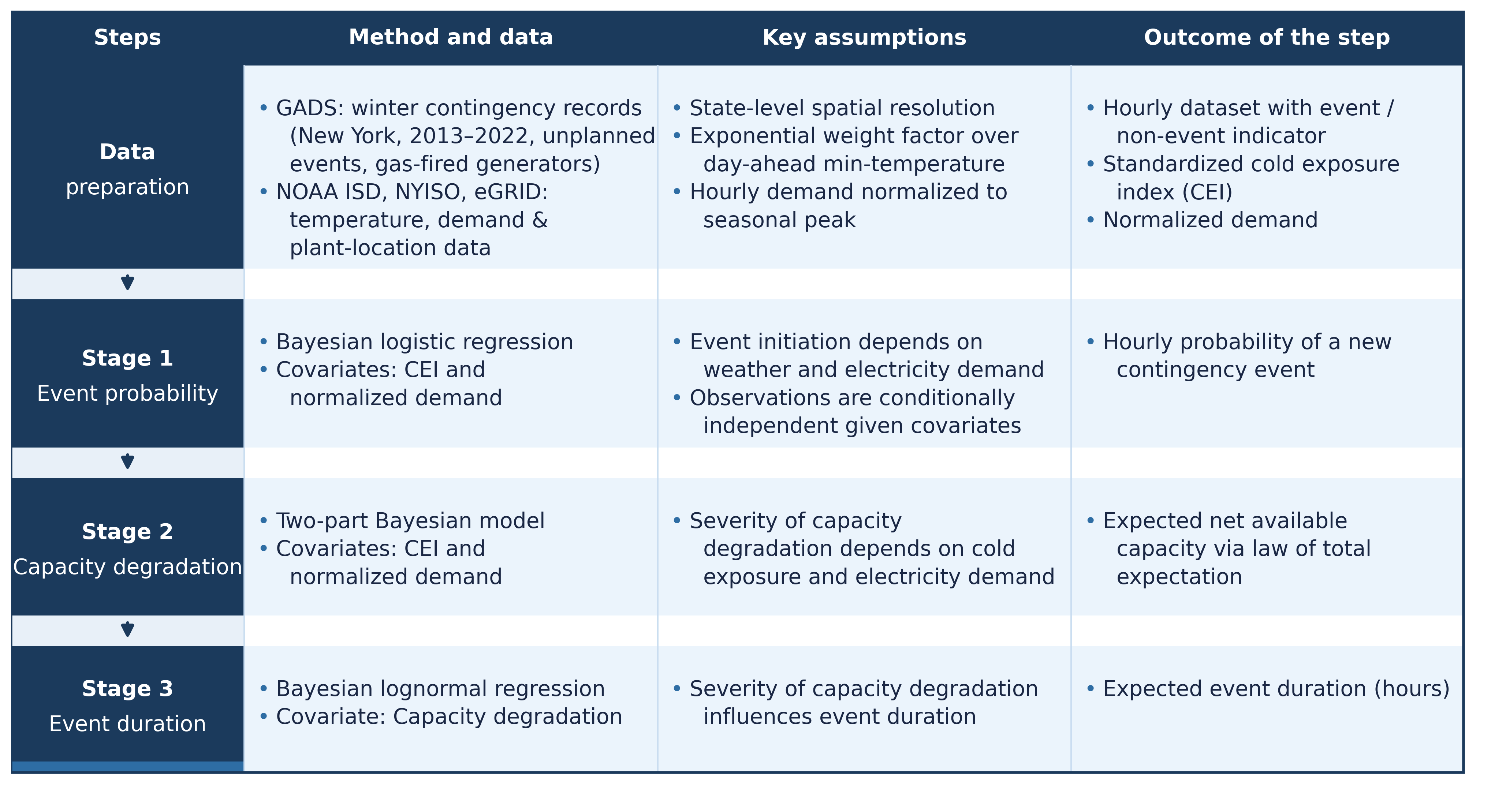}
\caption{Overview of the three-stage probabilistic framework for winter-driven gas-fired generator contingency events.}
\label{fig:model_overview}
\end{figure*}

Fig.~\ref{fig:model_overview} provides an overview of the proposed framework. The three stages are applied sequentially. 
Stage~1 estimates the probability that
a contingency begins ($E_t=1$) given $\mathrm{CEI}$ and $\mathrm{D_\mathrm{norm}}$. Conditional on
$E_t=1$, Stage~2 estimates the distribution
of normalized net available capacity, including
the probability of a full outage and the
conditional capacity distribution for
$\mathrm{NAC}_{\mathrm{norm}}>0$. Stage~3 then
estimates the duration distribution conditional
on the capacity outcome. During model fitting,
this capacity outcome is the observed
$\mathrm{NAC}_{\mathrm{norm}}$ recorded for
each event. For partial
derates, this dependence is represented through
the capacity loss
$\Delta_{\mathrm{cap}}=1-\mathrm{NAC}_{\mathrm{norm}}$.
In the present specification, duration has no
additional dependence on $\mathrm{CEI}$ or $\mathrm{D}_{\mathrm{norm}}$
once event type and capacity loss are given.
Each stage is therefore fitted separately
using the observations available for its
conditional model.

This conditional formulation is particularly useful for the present application because directly estimating the full joint distribution of initiation, capacity loss, and duration would require substantially more data across all combinations of operating conditions and event outcomes. Contingencies are relatively sparse compared with normal operating hours, and severe derates and long-duration events are sparser still. A direct joint model would therefore involve a high-dimensional and poorly populated outcome space. The sequential formulation reduces this estimation complexity by allowing each component to be learned from the subset of observations that is informative for that component, while retaining dependencies through conditioning.

The decomposition also follows the structure of the available data. Event initiation is estimated from hourly records containing both event and non-event hours (Section~\ref{sec2_hourlydata}), whereas available capacity and duration can only be observed and estimated from event-level GADS records after a contingency has occurred. The conditional formulation therefore aligns the statistical model with how the corresponding outcomes are actually observed.

An additional advantage of this representation is that it naturally supports sequential grid simulation. A simulated trajectory can first sample whether a contingency occurs, then sample the associated capacity reduction conditional on that occurrence, and finally sample how long that realized reduction persists. In this way, the framework provides a probabilistically coherent representation of contingency occurrence, severity, and persistence without requiring direct estimation of a single high-dimensional joint distribution.

The Bayesian framework is motivated by the
limited number of winter-driven contingency
events, which provides sparse information
for estimating model parameters. Under limited-data conditions, conventional point-estimation approaches can produce parameter estimates with substantial sampling variability, while standard confidence intervals based on large sample assumptions may be less reliable. Bayesian estimation describes
parameter uncertainty through posterior
distributions conditional on the observed
data, priors, and model assumptions. It yields a posterior distribution for each parameter, allowing both the most plausible parameter values and the degree of uncertainty associated with them to be quantified. 
Each stage is fitted separately using MCMC
to approximate these distributions.
Posterior draws are then used to calculate
event probabilities, expected net available
capacity, and expected durations, together
with credible intervals. This allows the
reported estimates to reflect uncertainty
arising from the limited evidence available
for winter-driven contingencies.
 
\subsection{Stage~1: Hourly event probability}\label{sec2_s1}

Stage~1 estimates the probability that a new winter-driven contingency event is
initiated during any given hour as a function of prevailing conditions.
Let $\mathbf{x}_t = (x_{1,t},\, x_{2,t},\, \ldots,\, x_{k,t})$ denote the
covariate vector at hour $t$. The binary event indicator is modeled as $E_t \mid \mathbf{x}_t \sim \mathrm{Bernoulli}(p_t)$, where
\begin{equation}
  p_t = \Pr(E_t = 1 \mid \mathbf{x}_t) = \sigma\!\left(\beta_0 + \sum_{j=1}^{k} \beta_j\, x_{j,t}\right)
  \label{eq:s1_likelihood}
\end{equation}
Here, $\sigma(\cdot)$ is the logistic function, $\beta_0$ is the intercept, and $\beta_j$ are the covariate coefficients. Weakly informative normal priors are specified for all parameters. In this study $k = 2$, with $x_1 = \mathrm{CEI}$ and 
$x_2 = D_{\mathrm{norm}}$, so the coefficients are denoted 
$\beta_{\mathrm{CEI}} \equiv \beta_1$ and 
$\beta_{D_{\mathrm{norm}}} \equiv \beta_2$ throughout.

The Bernoulli likelihood is adopted because the hourly contingency onset is strictly binary: $E_t = 1$ when an event is initiated and $0$ otherwise. The logistic function guarantees that predicted probabilities remain bounded within $[0, 1]$ while allowing failure likelihood to vary smoothly with cold exposure and system demand. We assume that these variables enter linearly on the log-odds scale, providing a simple model with few parameters given the limited number of event-initiation hours. This formulation is consistent with prior power system literature that uses logistic regression to model temperature- and load-dependent generator outage transitions \cite{murphy2019time}. The model describes how event-initiation probability varies with the available weather and demand inputs; it does not represent individual equipment failure mechanisms (which are unobservable in aggregated dataset).

\subsection{Stage~2: Conditional estimation of net available capacity}\label{sec2_s2}

Given that an event is initiated ($E_t = 1$), Stage~2 quantifies its capacity
impact through a two-part Bayesian model. Events are first classified as full outage ($\mathrm{NAC}_{\mathrm{norm}} = 0$) or derating ($0<\mathrm{NAC}_{\mathrm{norm}}<1$) events; the normalized available capacity is then modeled conditionally for partial derating.
A two-part specification is used because
$\mathrm{NAC}_{\mathrm{norm}}$ is bounded on $[0,1)$ and has a point mass at
zero: a full outage removes the generating unit entirely, whereas a derating event leaves
it operational at reduced output. A single continuous regression cannot
represent a discrete point mass and a continuous distribution simultaneously. Following the same reasoning as in Stage~1, the logistic component captures the categorical response and models the probability of a partial derate rather than a full outage. For partial derates, linear regression provides an interpretable first-order approximation of how derated capacity varies with CEI and demand given the limited sample.

In the first part, the probability that an event results in a derate rather than a full outage is modeled as a Bayesian logistic regression on $\mathbf{x}_t$:
\begin{equation}
  \Pr\!\left(\mathrm{NAC}_{\mathrm{norm}} > 0 \mid \mathbf{x}_t\right)
  = \sigma\!\left(\gamma_0 + \sum_{j=1}^{k} \gamma_j\, x_{j,t}\right)
  \label{eq:s2_logit}
\end{equation}
where $\gamma_0$ is an intercept and $\gamma_j$ are covariate coefficients.
The complement,
$1-\Pr(\mathrm{NAC}_{\mathrm{norm}}>0\mid\mathbf{x}_t)$, is the probability
that the event is a full outage.

In the second part, conditional on a derate, $\mathrm{NAC}_{\mathrm{norm}}$ is modeled with a Bayesian linear regression fitted exclusively on derate observations:
\begin{equation}
  \mathbb{E}\!\left[\mathrm{NAC}_{\mathrm{norm}} \mid \mathrm{NAC}_{\mathrm{norm}} > 0,\,\mathbf{x}_t\right]
  \;=\;
  \delta_0 + \sum_{j=1}^{k} \delta_j\, x_{j,t}
  \label{eq:s2_linear}
\end{equation}
where, $\delta_0$ is the intercept, $\delta_j$ are covariate slopes.

The final expected normalized net available capacity conditional on an event occurring is computed using the law of total expectation~\cite{belotti2015twopm}. 
\begin{equation}
  \mathbb{E}\!\left[\mathrm{NAC}_{\mathrm{norm}} \mid \mathbf{x}_t\right]
  = \Pr\!\left(\mathrm{NAC}_{\mathrm{norm}} > 0 \mid \mathbf{x}_t\right)
    \times
    \mathbb{E}\!\left[\mathrm{NAC}_{\mathrm{norm}}
                      \mid \mathrm{NAC}_{\mathrm{norm}} > 0,\,\mathbf{x}_t\right].
  \label{eq:s2_combined}
\end{equation}
Similar to Stage~1, $k = 2$ with $x_1 = \mathrm{CEI}$ and 
$x_2 = D_{\mathrm{norm}}$ in Stage~2, giving $\gamma_{\mathrm{CEI}}$, 
$\gamma_{D_{\mathrm{norm}}}$, $\delta_{\mathrm{CEI}}$, and 
$\delta_{D_{\mathrm{norm}}}$ in our current study. However, the framework is not restricted to these variables. With richer data, additional covariates such as fuel supply constraints, generator-specific characteristics, or higher-resolution weather inputs can be incorporated within the same formulation.

The logistic component uses all recorded events,
while the linear component uses only observations
with $\mathrm{NAC}_{\mathrm{norm}}>0$.
The likelihood separates into two components with
distinct parameter sets. With independent priors
between these components and no shared parameters,
their posterior distributions factorize, allowing
the two components to be estimated separately. Parameter uncertainty is
propagated into Eq.~\eqref{eq:s2_combined} by pairing
independent posterior draws from the two components.
For each pair, the predicted probability of
$\mathrm{NAC}_{\mathrm{norm}}>0$ is multiplied by
the conditional expected normalized net available
capacity, evaluated at the same covariate values.
The resulting values are summarized by their mean
and 95\% credible interval (CI), representing uncertainty
in the expected normalized net available capacity
conditional on an event occurring.

\subsection{Stage~3: Duration of contingency events}\label{sec2_s3}

Stage~3 characterizes the duration of winter-driven contingency events as a function of their severity. Event duration is defined as the elapsed time between the start and end of each GADS event. Because full outage and partial derating events exhibit distinct severity profiles, durations are modeled separately for the two classes. 

For full outages ($\mathrm{NAC}_{\mathrm{norm}} = 0$), duration ($\tau_{\mathrm{out}}$) is modeled using an intercept-only Bayesian lognormal distribution and the expected duration 
($\mathbb{E}[\tau_{\mathrm{out}}]$) for full outage events follows directly from the lognormal transformation. 
\begin{equation}
\begin{aligned}
  \log(\tau_{\mathrm{out}}) &\sim \mathcal{N}\!\left(\theta_{0}^{\mathrm{out}},\;\sigma_{\mathrm{out}}^2\right) \\[4pt]
  \mathbb{E}[\tau_{\mathrm{out}}]
  &= \exp\!\left(\theta_{0}^{\mathrm{out}} + \frac{\sigma_{\mathrm{out}}^2}{2}\right)
\end{aligned}
\label{eq:s3_outage}
\end{equation}

Here, $\theta_0^{\mathrm{out}}$ is the mean
log-duration and $\sigma_{\mathrm{out}}$ is its
residual standard deviation. Since capacity loss
is fixed at unity for full outages, it cannot
explain differences in their durations. The
intercept-only specification therefore provides
a baseline duration distribution for this class.

For partial derating events ($0<\mathrm{NAC}_{\mathrm{norm}}<1$), the event severity is defined as the capacity reduction $\Delta_{\mathrm{cap}}$, and duration is modeled using $\Delta_{\mathrm{cap}}$ as the sole covariate:
\begin{equation}
\begin{aligned}
  \Delta_{\mathrm{cap}} &= 1 - \mathrm{NAC}_{\mathrm{norm}} \\[4pt]
  \log(\tau_{\mathrm{derate}}) &\sim \mathcal{N}\!\left(\theta_{0}^{\mathrm{der}} + \theta_{1}^{\mathrm{der}}\,\Delta_{\mathrm{cap}},\;\sigma_{\mathrm{der}}^2\right)
\end{aligned}
\label{eq:s3_derate}
\end{equation}
where $\tau_{\mathrm{derate}}$ is the event duration for partial derating events, $\theta_{0}^{\mathrm{der}}$ is an intercept, $\theta_{1}^{\mathrm{der}}$ captures the dependence on severity, and $\sigma_{\mathrm{der}}$ is the residual standard deviation. 

The expected duration for partial derating events at a given severity is obtained by transforming the lognormal distribution:
\begin{equation}
  \mathbb{E}[\tau_{\mathrm{derate}} \mid \Delta_{\mathrm{cap}}]
  = \exp\!\left(\theta_{0}^{\mathrm{der}} + \theta_{1}^{\mathrm{der}}\,\Delta_{\mathrm{cap}} + \frac{\sigma_{\mathrm{der}}^2}{2}\right)
  \label{eq:s3_mean}
\end{equation}
 
 The lognormal specification is chosen because event durations are strictly positive and their observed distribution is right-skewed. Full outages and partial derates are modeled separately to allow their duration distributions to differ. For full outages, capacity loss is fixed at unity and therefore cannot explain differences in duration. We use an intercept-only model as a baseline. For partial derates, capacity loss varies across events and enters linearly in the mean log-duration. This provides a simple way to examine whether more severe derates are associated with longer events, without assuming that capacity loss alone determines event duration. In reality, event duration may reflect several factors, including fault diagnosis, crew and parts availability, equipment repair, and restoration of fuel supply. These factors may contribute to the observed variation and long durations, but the available records do not separate their contributions. We therefore use the lognormal distribution as a statistical description of positive, right-skewed durations rather than as a distribution derived from a specific restoration mechanism. Incorporating covariates such as failure mechanism, operational response (e.g., crew dispatch or repair actions), or higher-resolution system conditions may improve the accuracy of duration estimates.

\section{Case Study: New York State}

\begin{figure*}[t!]
\centering
\includegraphics[width=\textwidth]{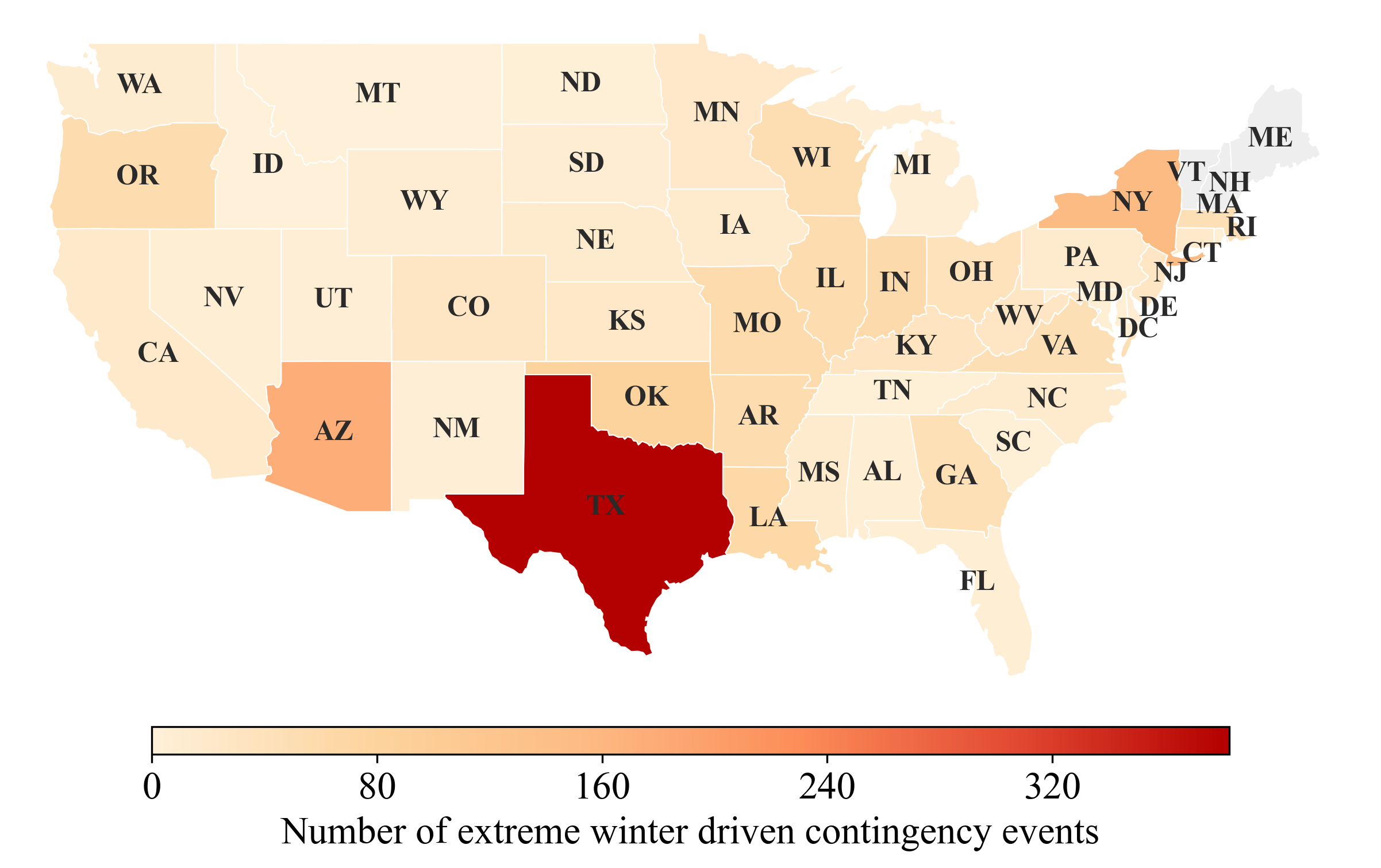}
\caption{
State-level counts of winter-weather-related gas-fired generator contingency events identified using structured filtering from GADS records (2013--2022).
}
\label{fig:us_winter_map}
\end{figure*}

The filtering method described in Section~\ref{sec2_gadsdata} yields 1,779 winter-driven contingency events across 47 U.S. states between 2013 and 2022. Of the 1,779 events, 1,060 (59.6\%) are full outages and 719 (40.4\%) are partial derating. The resulting pattern is semi-continuous, with a large point mass at zero and a broad distribution of positive derating levels. The geographic distribution (Fig.~\ref{fig:us_winter_map}) shows that these disruptions extend across multiple climatic zones rather than being limited to the coldest parts of the country. For example, Texas exhibits the highest event concentration, indicating that gas-fired generation can be highly vulnerable to extreme cold even in regions where winterization has historically not been considered a priority.

\begin{figure}[h!]
\centering
\includegraphics[width=\columnwidth]{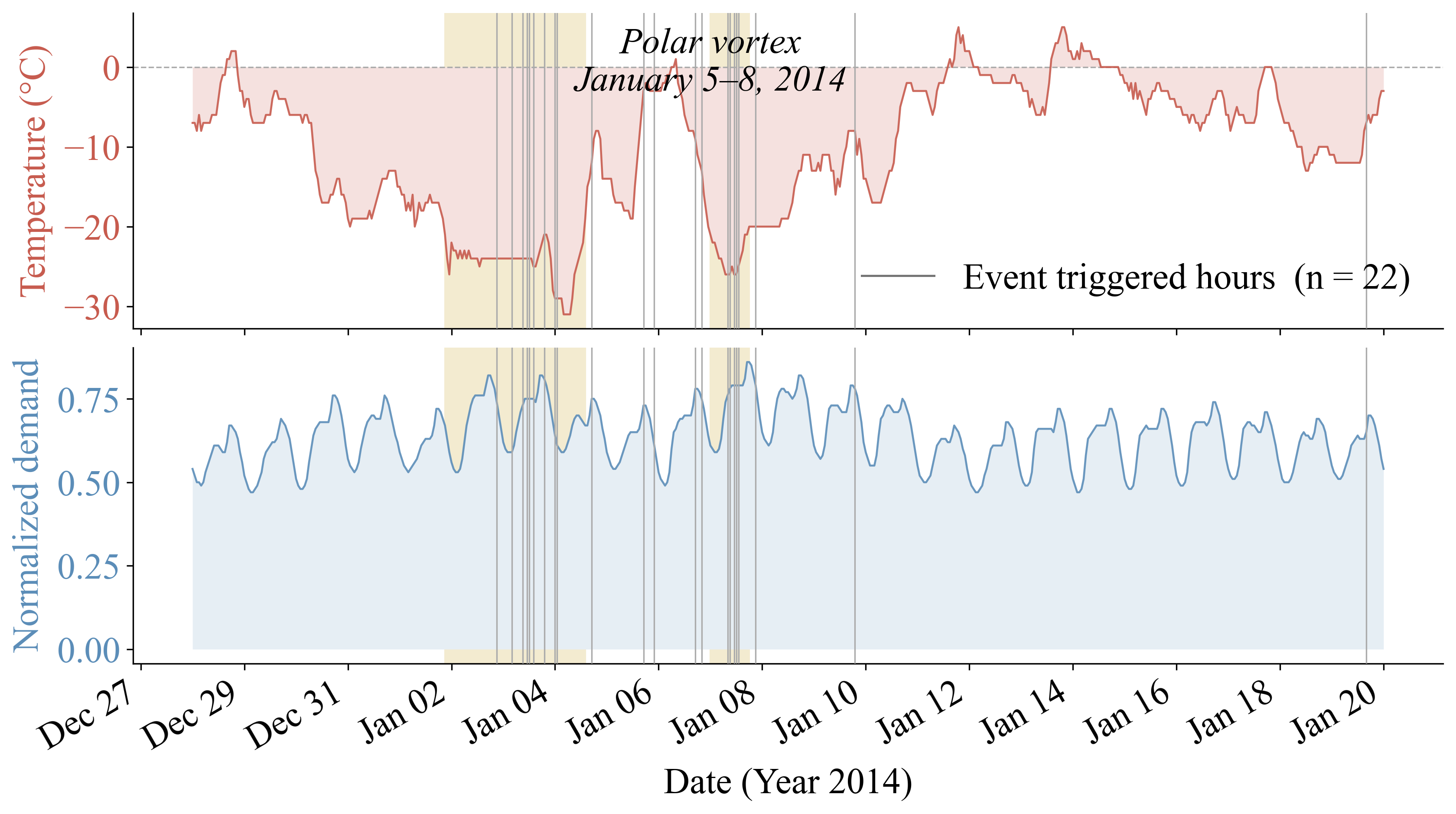}
\caption{Winter conditions and generator contingency events in New York during the 2014 polar vortex.}
\label{fig:ny_timeseries}
\end{figure}

However, due to data limitation we are constrained to rely on aggregated temperature and demand data. For large, geographically diverse states, such aggregation can obscure more than it reveals. Hence we choose New York state (NY) as our test case. NY experiences recurring subfreezing winter conditions, has a single system operator (NYISO) with publicly available hourly electricity demand data, and is geographically compact enough that a state-level temperature signal remains informative. Fig.~\ref{fig:ny_timeseries} shows a representative excerpt of the constructed hourly dataset for NY, with observed state-level temperature, normalized demand, and generator contingency event initiations during the 2014 polar vortex~\cite{NERC2014PolarVortex}. From the data, it is observed that generator contingency events tend to occur during periods of prolonged cold exposure with elevated electricity demand. This pattern supports the central hypothesis of this work that extreme cold temperature and elevated electricity demand jointly contribute to generator performance degradation and contingency risk.

\subsection{Data Patterns in New York}\label{sec3_patterns}
\begin{figure}[htbp]
\centering
\includegraphics[width=\columnwidth]{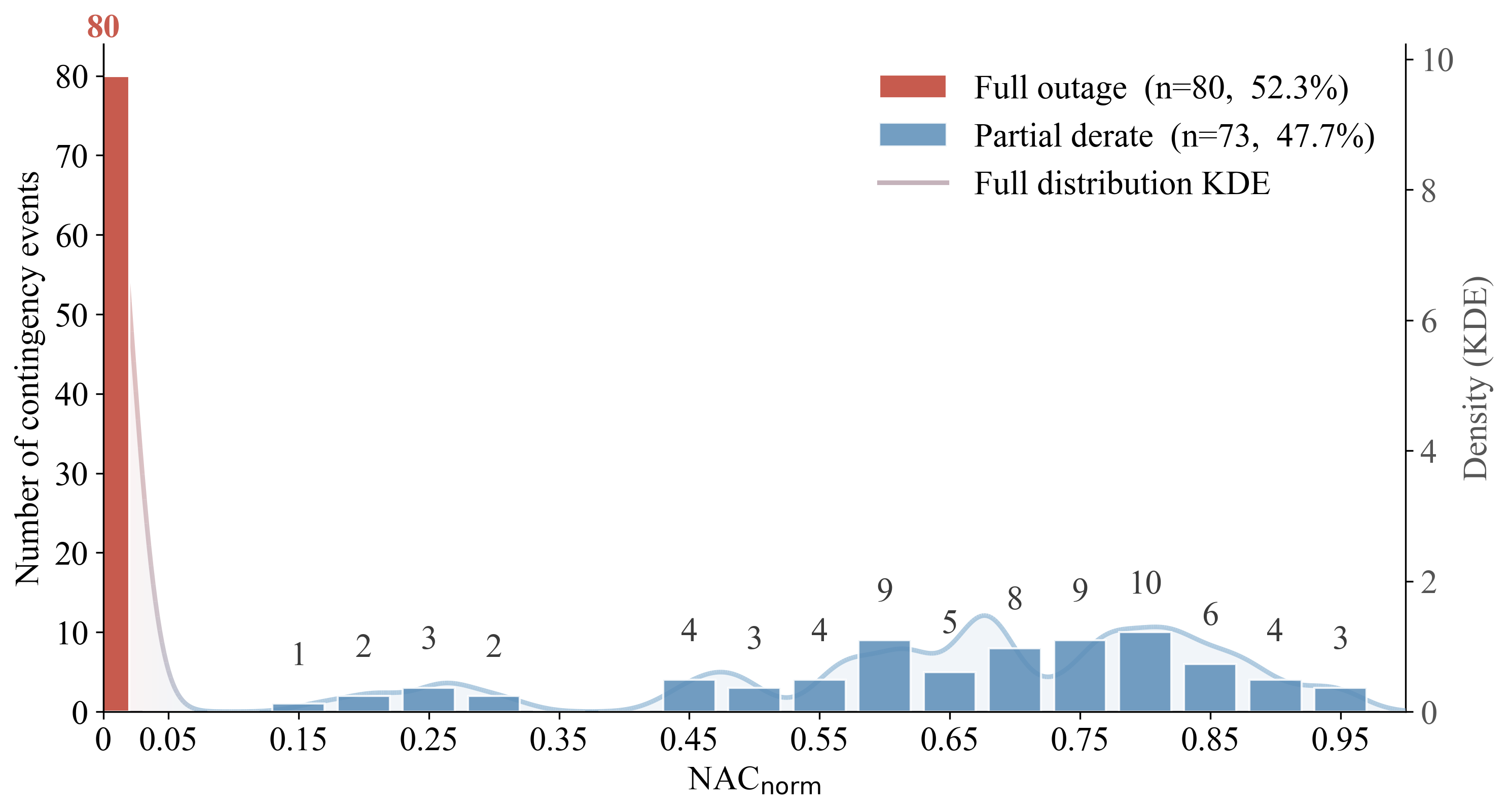}
\caption[Distribution of normalized net available capacity ($\mathrm{NAC}_{\mathrm{perc}}$) for winter-driven contingency events in New York.]{
Distribution of normalized net available capacity ($\mathrm{NAC}_{\mathrm{norm}}$) for winter-driven contingency events in New York.}
\label{fig:ny_nac_dist}
\end{figure}

Applying the same filtering procedure as in Section~\ref{sec2_gadsdata}, we identify 153 winter-driven gas-fired generator contingency events in New York between 2013 and 2022. 80 of these events (52.3\%) are full outages and 73 (47.7\%) are partial deratings, preserving the semi-continuous severity structure observed at the national level. The distribution of NY events is shown in Fig.~\ref{fig:ny_nac_dist}.

\begin{figure}[htbp]
\centering
\includegraphics[width=\columnwidth]{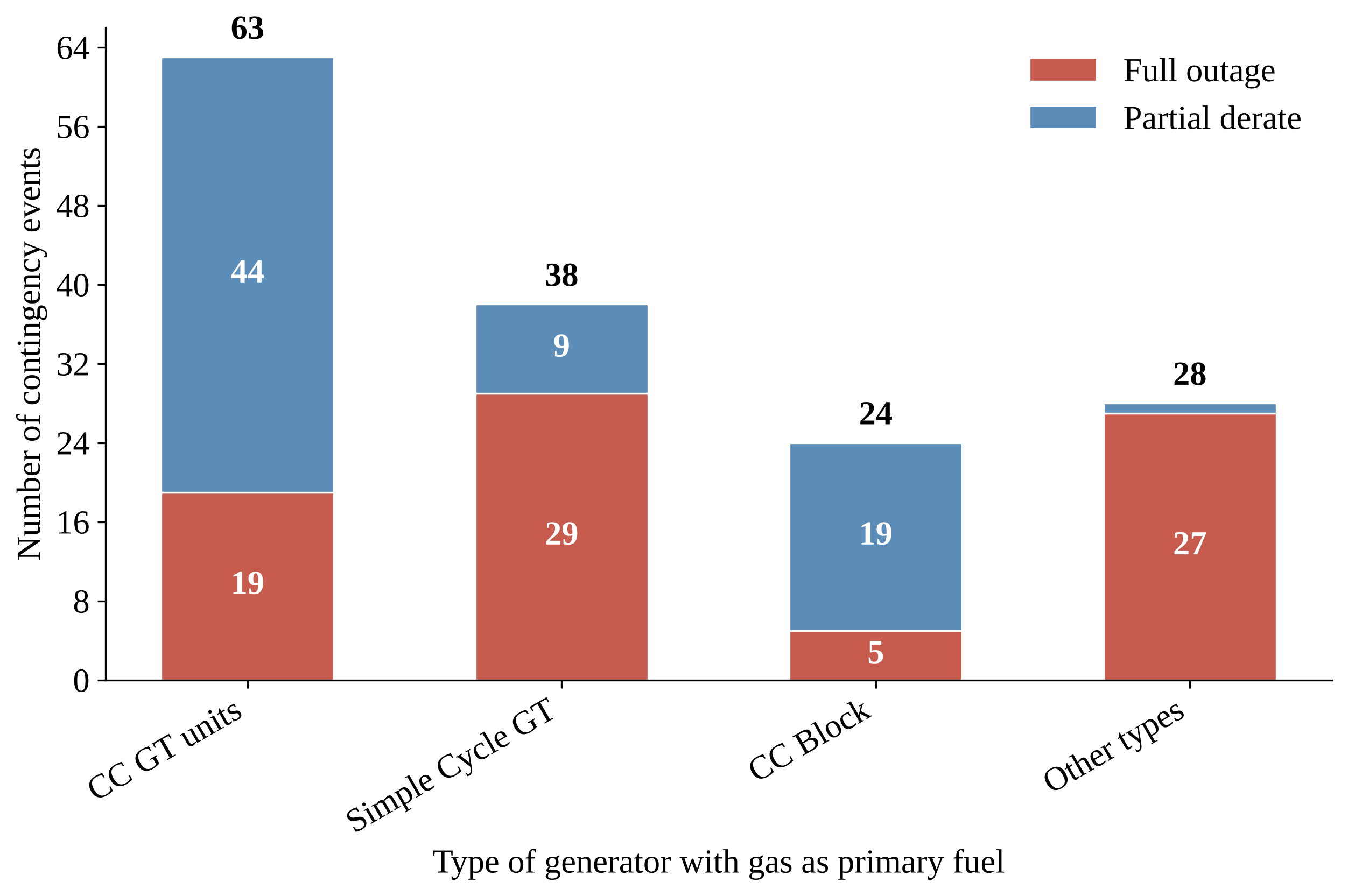}
\caption{Winter-driven gas-fired contingency events in New York (2013--2022) by generator unit type, split between full outages and partial derates}
\label{fig:ny_unit_type}
\end{figure}

We also observe distinct winter failure patterns when considering generator technology. Fig.~\ref{fig:ny_unit_type} shows different winter-driven events based on the generator technology. Combined cycle gas turbines account for the largest share of events (63), followed by simple cycle gas turbine units (38). The outage-to-derate balance differs across technologies: simple cycle units exhibit a higher fraction of full outages, whereas combined cycle units are more often observed to continue operating at reduced output. 

\begin{figure}[h!]
\centering
\includegraphics[width=0.6\columnwidth]{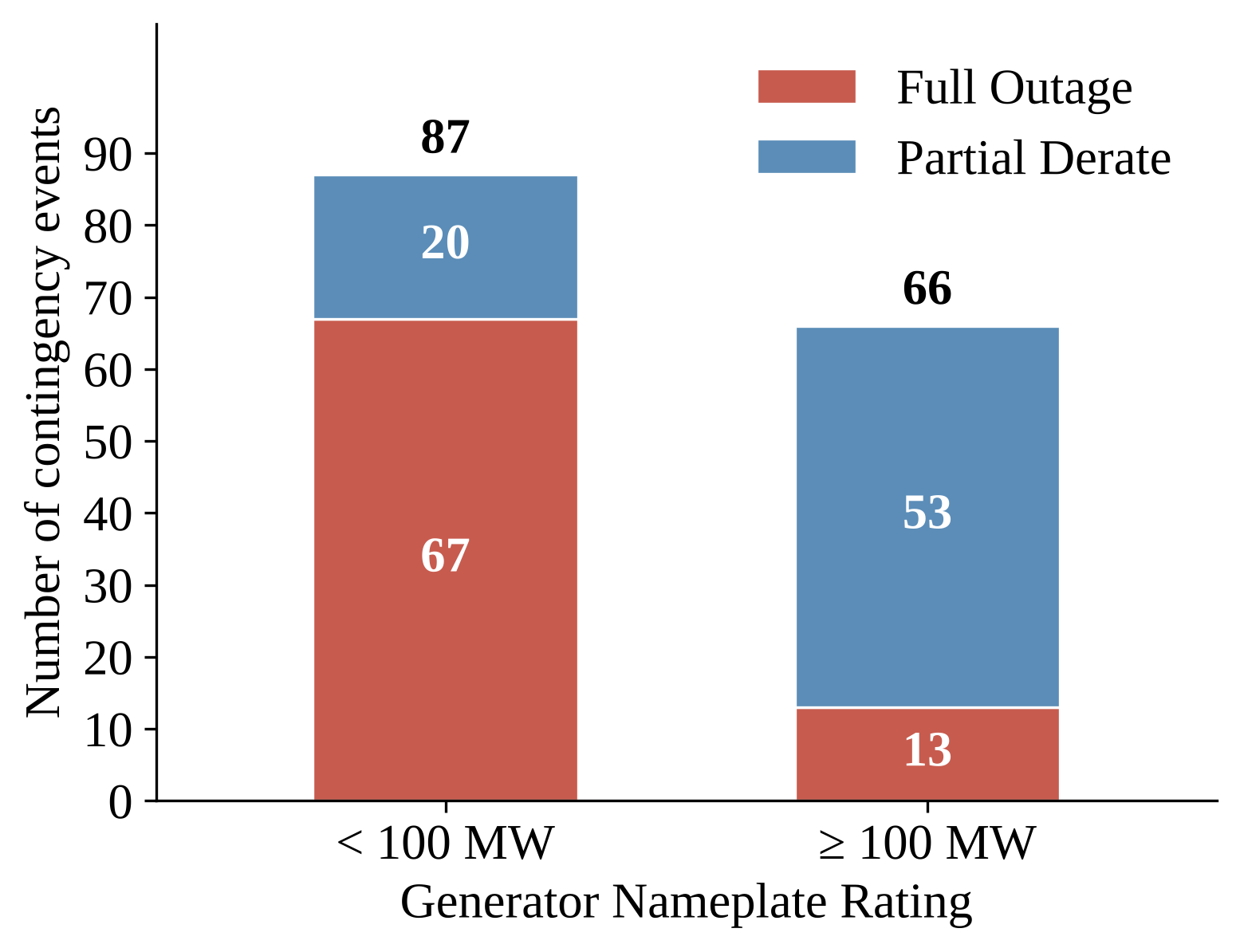}
\caption{Winter-driven contingency events in New York grouped by generator nameplate rating.}
\label{fig:ny_rating}
\end{figure}

A similar structural pattern appears across generator size classes as shown in Fig.~\ref{fig:ny_rating}. Units below 100 MW account for the largest number of events (87), followed by units above 100 MW (66). The fraction of full outages decreases with unit size while the share of partial derating increases, indicating that larger units are more likely to remain online at reduced capacity rather than fail completely during extreme winter conditions. Available GADS report generator size in discrete rating groups rather than exact nameplate capacities, so these grouped categories are used throughout the analysis. Even with this limitation, the observed pattern suggests systematic differences in extreme winter response across generator scales.

The limited number of total NY events (153) within the study period constrains further disaggregation the model training data by unit type or size without substantially reducing sample size. Hence, we model generator behavior at an aggregated level while capturing uncertainty explicitly through a Bayesian framework, which is well suited to rare events and limited data settings. 

\subsection{Selected weather stations}\label{sec3_stations}
\begin{figure}[htbp]
\centering
\begin{subfigure}{\textwidth}
    \centering
    \includegraphics[width=0.8\textwidth]{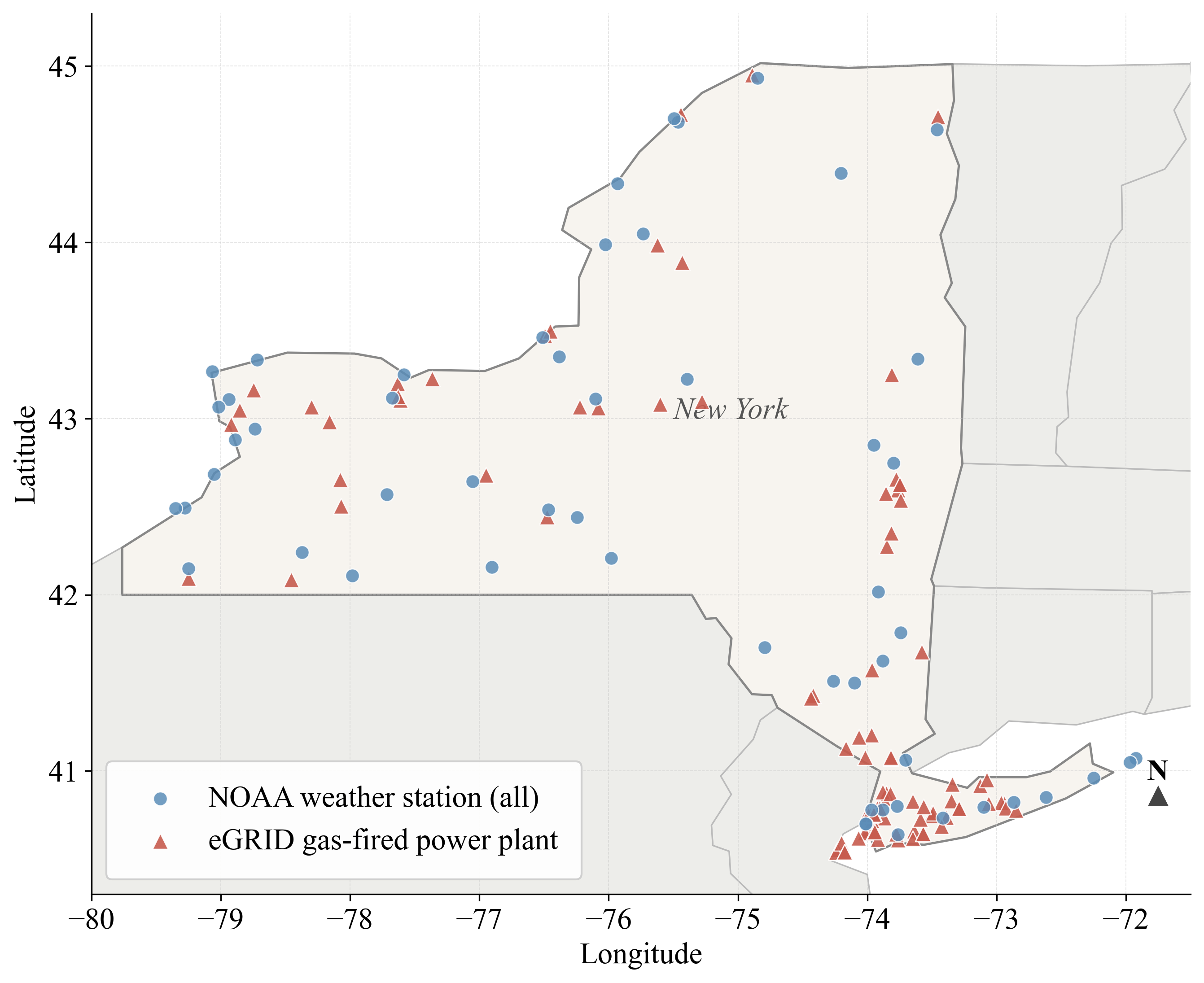}
    \caption{}
    \label{fig:ny_stations_before}
\end{subfigure}
\vspace{0.3cm}
\begin{subfigure}{\textwidth}
    \centering
    \includegraphics[width=0.8\textwidth]{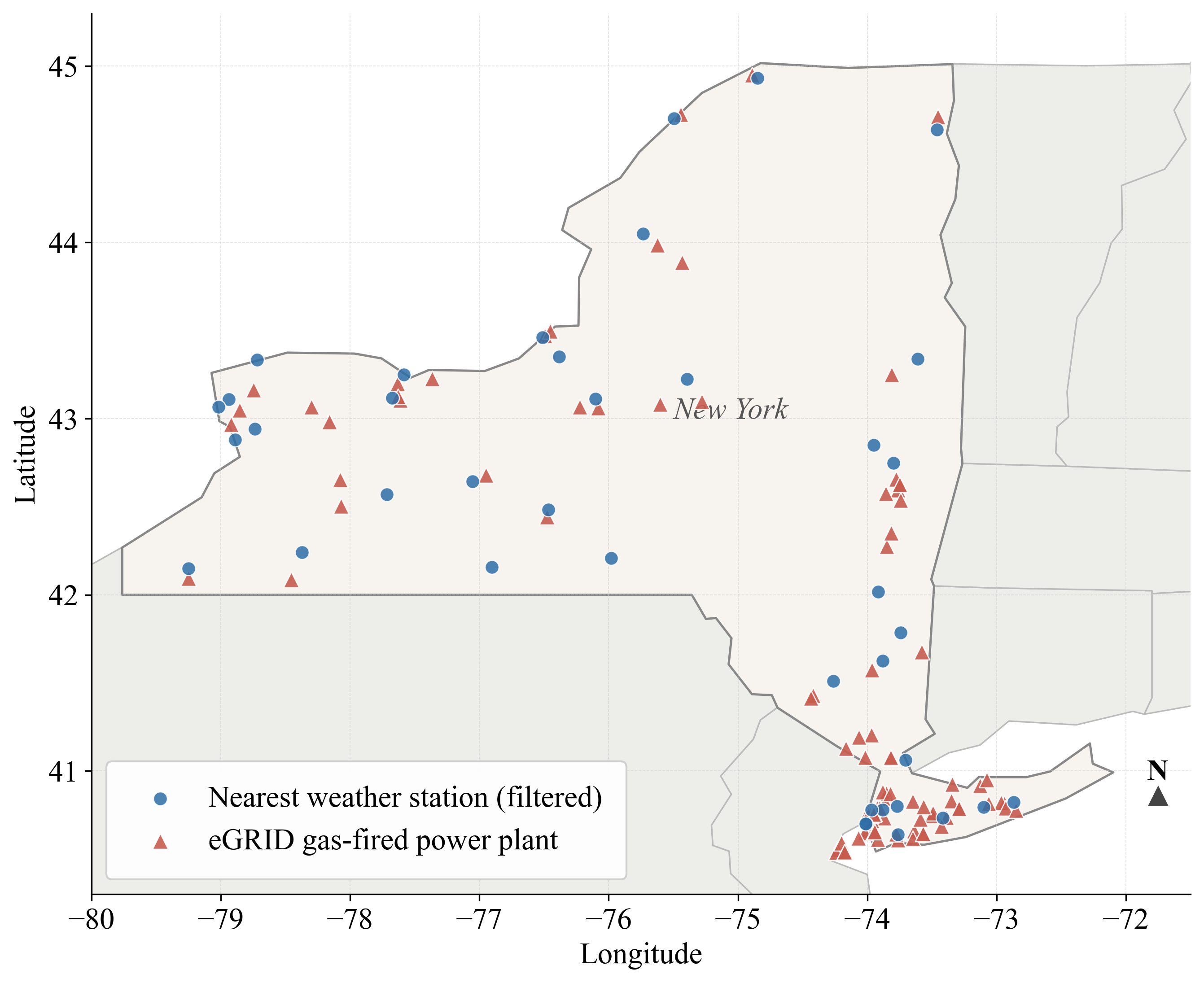}
    \caption{}
    \label{fig:ny_stations_after}
\end{subfigure}
\caption{Selection of representative weather stations in New York (a) before filtering, (b) after filtering.}
\label{fig:ny_station_selection}
\end{figure}

We use eGRID data to identify the relevant ISD weather stations for NY states. After applying the spatial filtering procedure described in Section~\ref{sec2_weatherdata}, the 55 NOAA ISD stations within NY are reduced to 39 stations that are representative of the 106 gas-fired power plants locations identified from eGRID 2022. Fig.~\ref{fig:ny_stations_before} and Fig.~\ref{fig:ny_stations_after} show the station and plant locations before and after filtering.

\subsection{CEI and decay parameter}\label{sec3_params}
\begin{figure}[htbp]
\centering
\includegraphics[width=\linewidth]{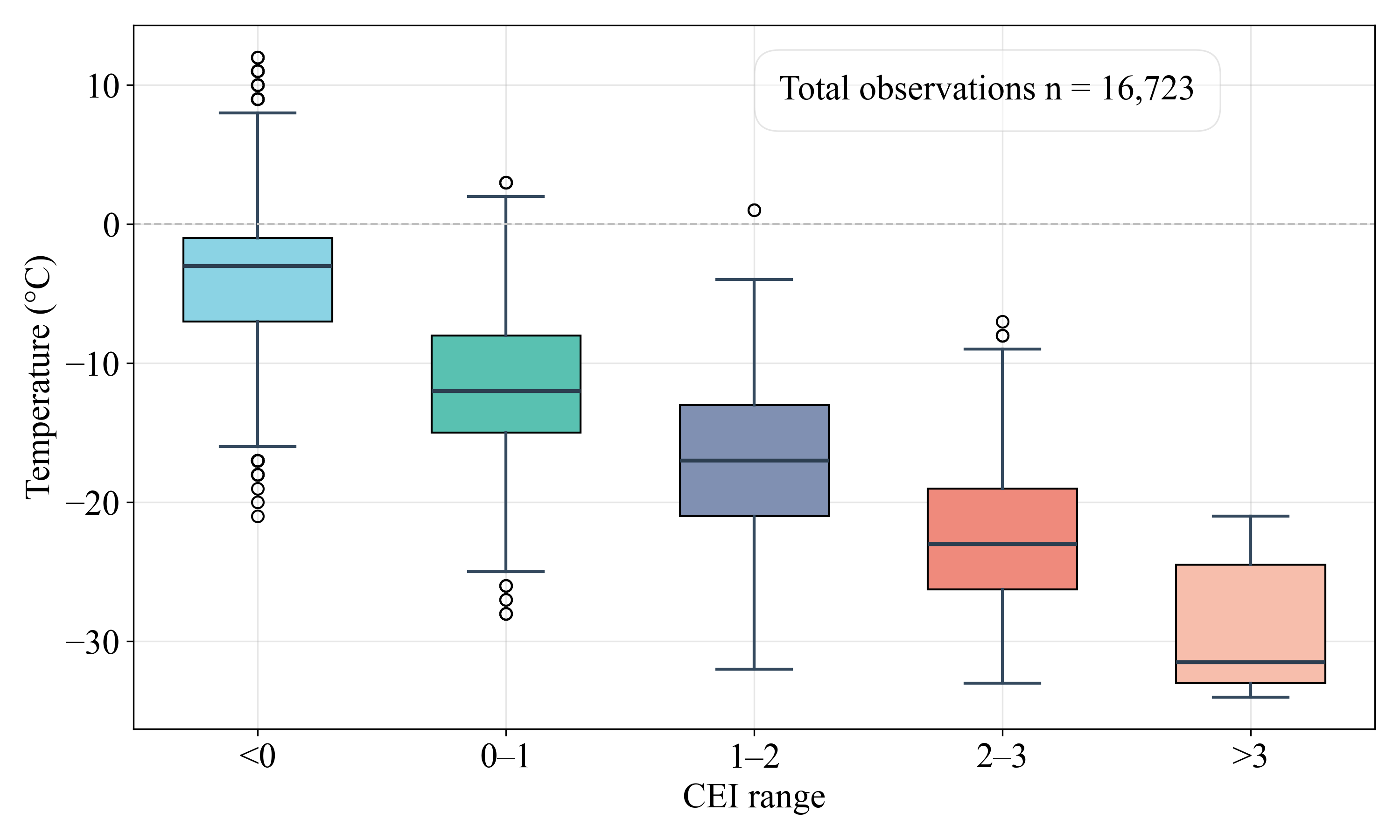}
\caption{Distribution of day-ahead temperatures across ranges of the cold-exposure index (CEI).}
\label{fig:cei_temp_box}
\end{figure}

The decay parameter $\lambda$ in Eq.~\eqref{eq:cei_raw} controls the effective temporal memory of cold exposure. To determine an appropriate value, we estimated $\lambda$ directly from the data within a Bayesian formulation of the Stage~1 model. The posterior distribution was concentrated around $\lambda \approx 0.97$ (posterior mean $0.971$, standard deviation $0.020$), indicating that temperatures from up to 24 hours prior retain a substantial influence on current system conditions. Under this value, temperatures 24 hours earlier retain approximately $0.97^{24} \approx 48\%$ of the weight of the current hour. A sensitivity analysis across candidate values $\lambda \in [0.70, 0.99]$ confirms that the predictive performance varies only marginally for $\lambda$ in the range $0.95$--$0.99$. Based on these results, we  fix $\lambda = 0.97$ in the definition of $\mathrm{CEI}$ throughout the analysis. The $CEI$ standardization is performed using training-data statistics, with $\mu^{\mathrm{train}}_{\mathrm{CEI}} = 152.02$ and $\sigma^{\mathrm{train}}_{\mathrm{CEI}} = 113.22$. To provide a physical interpretation of the $\mathrm{CEI}$, Fig.~\ref{fig:cei_temp_box} shows the distribution of day-ahead temperatures associated with different ranges of computed CEI values. Higher CEI values correspond to
colder and more persistent sub-freezing temperature profiles, while
lower CEI values are associated with near-freezing or above-freezing
conditions. This confirms that the index meaningfully captures the
accumulated cold exposure driving winter-driven generator stress.

\subsection{Model estimation and implementation}
\label{sec3_implementation}

The three-stage models are implemented in Python using NumPyro.
Stage~1 uses the New York hourly dataset, whereas
Stages~2 and~3 use New York event-level records during the study period. Normal priors are used for the regression coefficients in each stage, with slope priors centered at zero. Parameter estimation is performed via Markov Chain Monte Carlo (MCMC) sampling using the No-U-Turn Sampler (NUTS). Each component is fitted with two chains, 1,000 warm-up
iterations, and 2,000 retained draws per chain.
The target acceptance probability is 0.95.
Sampling is assessed using $\hat{R}$
\cite{vehtari2021rank}, effective sample sizes,
and divergent transitions. Across all fits,
including cross-validation, the maximum reported
$\hat{R}$ was 1.01, the minimum bulk effective
sample size was 926, and no divergent transitions
were recorded. Model outputs are calculated for each posterior draw. In Stage~1, the logistic
transformation gives event-initiation probabilities.
In Stage~2, independent draws from the two
components are paired, and the partial-derate
probability is multiplied by the conditional
expected net available capacity. The pairing
is fixed across the covariate grid. In Stage~3,
expected durations are calculated using
\eqref{eq:s3_outage} and \eqref{eq:s3_mean}.
The posterior mean and the 2.5th and 97.5th
percentiles of these calculated quantities
provide the reported estimates and their
95\% credible intervals.

\section{Results}

\subsection{Stage 1: Probability of the initiation of contingency events}
\begin{figure*}[h!]
\centering
\includegraphics[width=\textwidth]{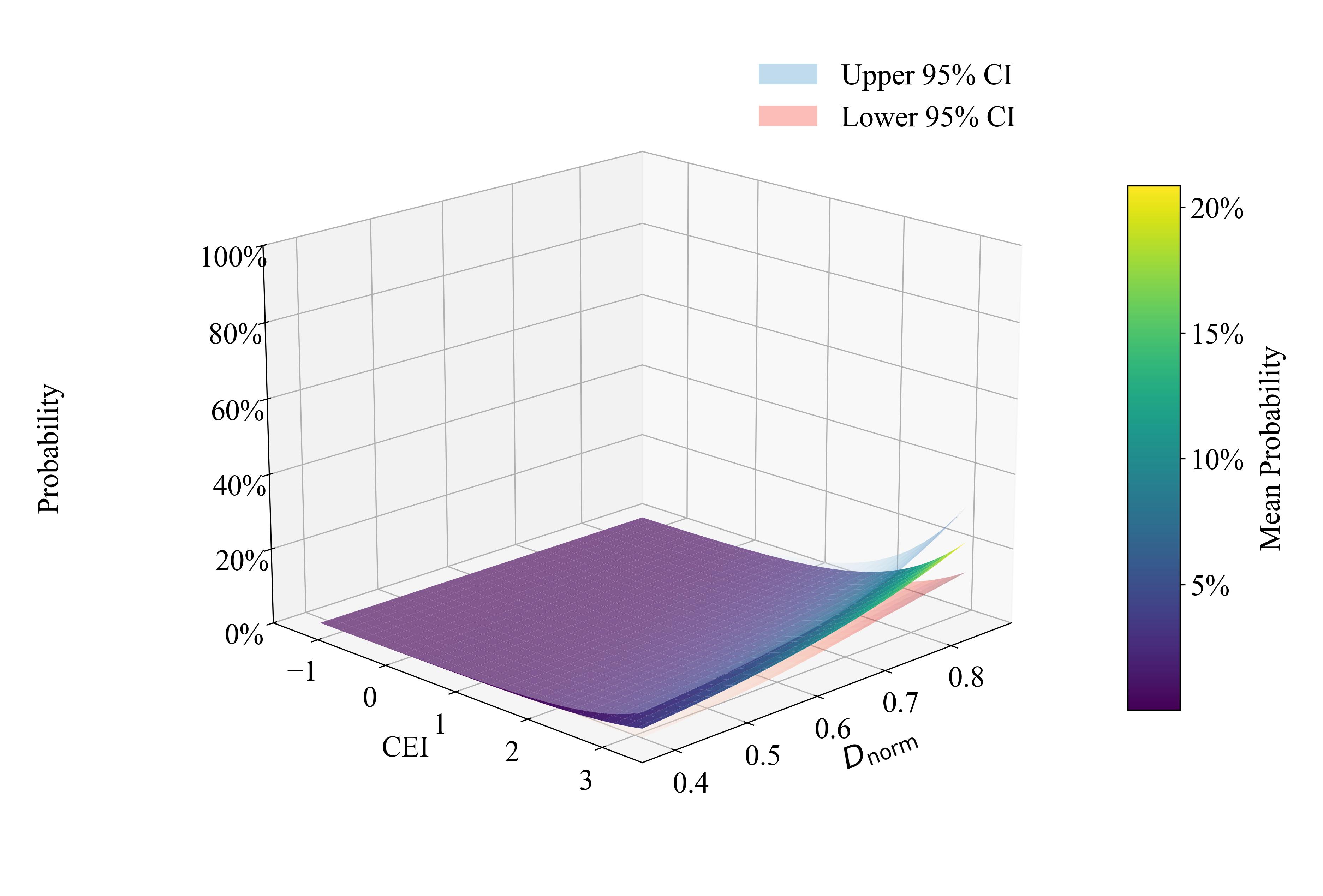}
\caption{
Estimated hourly probability of winter-driven gas-fired generator contingency events in NY.}

\label{fig:s1_prob_surface}
\end{figure*}

Fig.~\ref{fig:s1_prob_surface} shows the estimated hourly probability of a new winter-driven generator contingency event in New York as a function of $\mathrm{CEI}$ and $D_{\mathrm{norm}}$. This probability is learned from a systematically developed hourly dataset that includes both event and non-event periods, constructed from raw GADS records as described in Section~\ref{sec2_hourlydata}. 

The central surface height in
Fig.~\ref{fig:s1_prob_surface} represents the
posterior mean event-initiation probability.
Its color represents the same probability,
increasing from purple to yellow.
The translucent blue and red surfaces show
the upper and lower bounds of the pointwise
95\% credible interval, respectively.
The separation between these bounds describes
uncertainty in the estimated probability at
each combination of CEI and normalized demand.

The surface plot exhibits a clear gradient from the low-${\mathrm{CEI}}$, low-$D_{\mathrm{norm}}$ corner to the upper extreme. Under conditions with ${\mathrm{CEI}} < -0.34$, predicted hourly probabilities are effectively zero regardless of demand level. These values correspond to temperature profiles in which the day-ahead temperatures remain at or near $0^\circ$C, resulting in low accumulated freezing exposure. Across the central region of the observed covariate space ($\mathrm{CEI} \approx 0.96$, $D_{\mathrm{norm}} \approx 0.62$), the predicted probability remains around 1\%. At these CEI levels, the day-ahead temperature includes sustained periods below $0^\circ C$, although not continuously throughout the entire day. The hourly contingency probability first exceeds 2\% at ${\mathrm{CEI}} \approx 0.58$
where the temperature profile typically reflects extended sub-freezing conditions over a large fraction of the preceding 24 hours. It then crosses 5\% to 15\% at ${\mathrm{CEI}}$ values of approximately 1.58 to 2.73, respectively. These levels represent progressively more severe cold exposure, with average day-ahead temperatures around $-15^\circ C$ to $-20^\circ C$ with no break above freezing, provided demand is also elevated ($D_{\mathrm{norm}} \geq 0.49$, and 0.75, respectively). At the most extreme end of observed conditions ($\mathrm{CEI} = 3.19$, $D_{\mathrm{norm}} = 0.86$), characteristic of polar-vortex-type events with day ahead mean temperatures around $-29^\circ C$, the predicted mean probability reaches 24\% (with 95\% credible interval: 14\%–36\%). The posterior distribution shows that the increase in risk is driven by both environmental and operational stress. The posterior mean for the demand coefficient is $\beta_{D_{\mathrm{norm}}} = 5.30$ (95\,\% credible interval: 2.44--8.09), and for the cold exposure coefficient is $\beta_{{\mathrm{CEI}}} = 1.17$ (95\,\% credible interval: 0.93--1.39). The larger coefficient on demand indicates that system loading amplifies cold-induced risk. For instance, fixing ${\mathrm{CEI}}$ at its 75th-percentile value and increasing $D_{\mathrm{norm}}$ from 0.39 to 0.85 raises the predicted probability from 1\,\% to 7\,\%. Although the absolute probabilities remain small at the hourly level, this reflects the rarity of winter-driven contingency events rather than weak dependence on the covariates. During extended periods of cold and high demand, these hourly probabilities accumulate, leading to a substantial increase in overall risk.

\subsection{Stage 2: Estimated net available capacity}
\begin{figure*}[h!]
\centering
\includegraphics[width=1.05\textwidth]{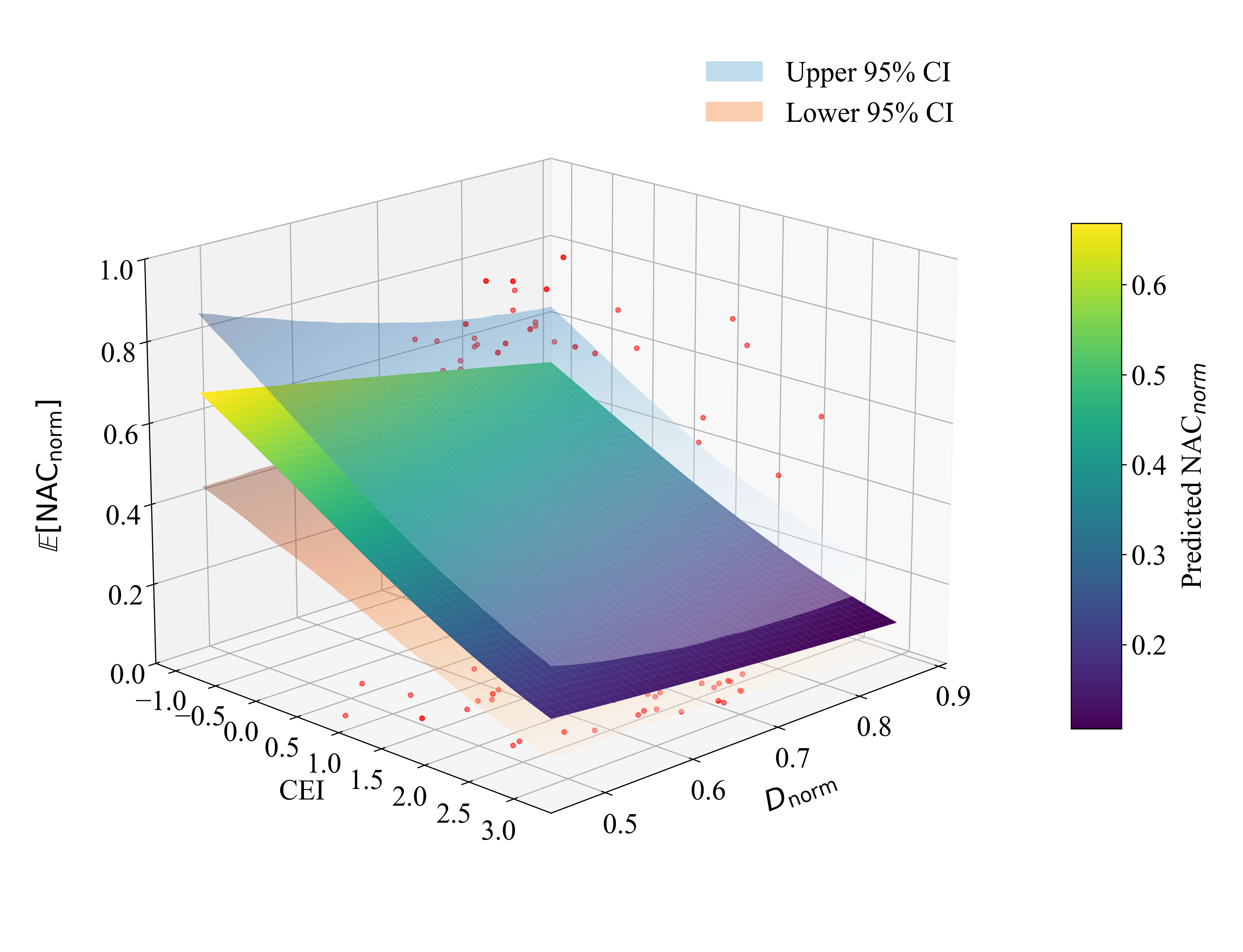}
\caption{Predicted normalized net available capacity ($\mathrm{NAC}_{\mathrm{norm}}$) of gas-fired generators in New York as a function of cold exposure index ($\mathrm{CEI}$) and normalized demand ($D_{\mathrm{norm}}$).}
\label{fig:s2_capacity_surface}
\end{figure*}

Conditional on a contingency event occurring, the two-part Bayesian model estimates both the type and severity of the resulting capacity loss as a joint function of scaled cold exposure (${\mathrm{CEI}}$) and normalized demand ($D_{\mathrm{norm}}$). The logistic component quantifies the probability that an event results in non-zero remaining capacity (i.e., partial derating), and the linear component estimates expected $\mathrm{NAC}_{\mathrm{norm}}$ conditional on a partial derating outcome occurring.  Applied to the 153 contingency events recorded in New York, Fig.~\ref{fig:s2_capacity_surface} shows the relationship between generator availability, cold exposure index ($\mathrm{CEI}$), and normalized demand ($D_{\mathrm{norm}}$), conditional on the occurrence of a contingency event. 
The central surface height in
Fig.~\ref{fig:s2_capacity_surface} represents
the posterior mean expected normalized net
available capacity conditional on an event
occurring. Its color represents the same
quantity, increasing from purple to yellow.
The translucent blue and orange surfaces show
the upper and lower bounds of the pointwise
95\% credible interval, respectively.
Red points represent observed event-level normalized
net available capacity. Observed events are expected to scatter more widely than these bounds, since the bounds describe uncertainty in the conditional mean
rather than the variability of observed data. The surfaces are
calculated using the same posterior-draw
pairs across the covariate grid as described in
Section~\ref{sec3_implementation}.

Under mild conditions (${\mathrm{CEI}} = -2.3$, $D_{\mathrm{norm}} = 0.50$), the posterior mean $\mathrm{NAC}_{\mathrm{norm}}$ is 0.66 (95\,\% credible interval: 0.39--0.93), indicating that, on average, affected generators retain two-thirds of their nameplate capacity. At the most severe observed stress conditions (${\mathrm{CEI}} = 3.19$, $D_{\mathrm{norm}} = 0.86$), 
the estimated mean $\mathrm{NAC}_{\mathrm{norm}}$ falls to 0.13 (95\,\% credible interval: 0.03--0.30). The lower bound of the credible interval approaches zero across much of the high-${\mathrm{CEI}}$, high-$D_{\mathrm{norm}}$ region, reflecting a substantial probability mass associated with full outage. At the parameter level (Table~\ref{tab:posteriors}), the CEI 
coefficient $\gamma_{\mathrm{CEI}}$ is clearly negative 
(mean $-0.577$, 95\,\% CI: $[-0.998, -0.175]$), indicating that 
greater cold exposure shifts events toward full outage. The demand 
coefficient $\gamma_{D_{\mathrm{norm}}}$ is also negative on average 
(mean $-0.773$) but its credible interval crosses zero 
($[-2.314, 0.691]$), reflecting limited precision in estimating the 
influence of system loading on the outage-versus-derate 
classification from the available data size. Taken together, these results show that winter stress affects not only the likelihood of generator failure, but also the severity of capacity loss among affected units.

\subsection{Stage 3: Estimated event duration}
\begin{figure}[htbp]
\centering
\includegraphics[width=0.8\columnwidth]{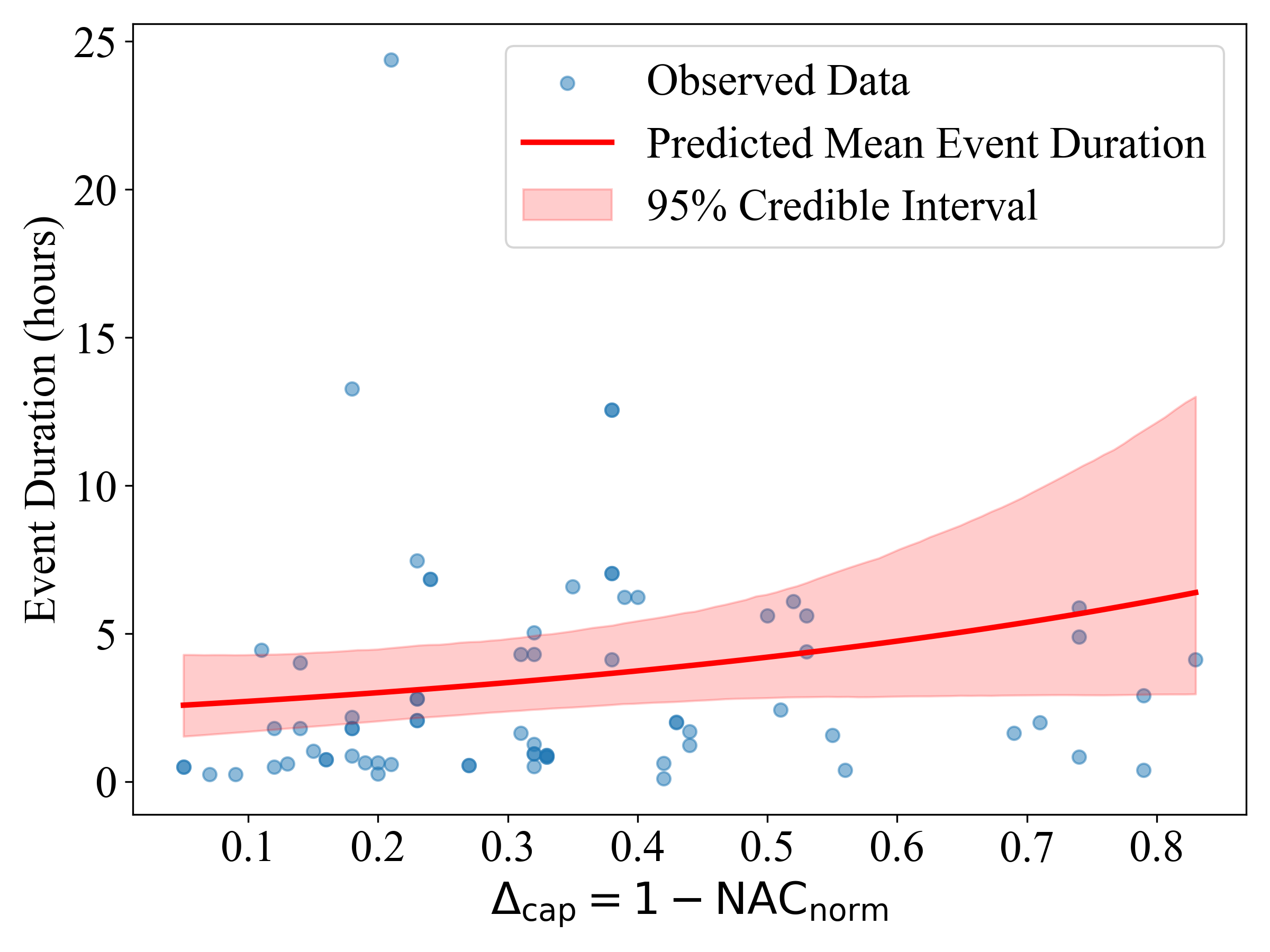}
\caption{
Event duration as a function of capacity loss for partial derating events.
}
\label{fig:duration_severity}
\end{figure}

Fig.~\ref{fig:duration_severity} shows event duration as a function of capacity loss for the 73 partial derating events in New York, modeled using a Bayesian lognormal regression with capacity loss as the sole covariate. The predicted event duration increases from 2.4 hours (95\,\% credible interval: 1.4--4.1\,h) at mild derates ($\Delta_{\mathrm{cap}} = 0.05$) to 7.1 hours (95\,\% credible interval: 2.6--18.3\,h) at the most severe derates ($\Delta_{\mathrm{cap}} = 0.83$). Credible intervals widen at higher capacity loss values, reflecting both data sparsity and increasing variability in event duration. For full outage events ($n = 80$), where $\Delta_{\mathrm{cap}}$ is fixed at unity, duration is modeled using an intercept-only lognormal formulation. The model estimates a typical event duration time of 12.7 hours (95\,\% credible interval: 8.0--21.2\,h), indicating substantially longer and more variable duration relative to partial derating events.

\subsection{Covariate dependence and parameter stability}
\label{sec3_dependence}
Because temperature and electricity demand
may contain overlapping information, we examine
dependence between cold exposure index ($\mathrm{CEI}$) and normalized demand ($D_{\mathrm{norm}}$) separately in the hourly and event-level datasets. We measure linear association using Pearson's correlation coefficient ($r$) and monotonic association using Spearman's rank correlation coefficient ($\rho$). In Stage~1 ($n=20,904$),
the association is moderately positive
($r=0.425$, $\rho=0.397$). The resulting variance inflation factor is $\mathrm{VIF} = 1.22$, calculated as $1/(1-r^2)$. This is far below conventional collinearity thresholds of concern ($\mathrm{VIF} > 5$ or $10$)~\cite{montgomery2021introduction}, suggesting limited linear collinearity between the inputs. In the Stage~2 event-level datasets, which condition on event occurrence to evaluate event classification and derated capacity, the association is even weaker, yielding $r = -0.190$, $\rho=-0.122$ and $r = -0.292$,  $\rho=-0.260$, for two parts respectively. The corresponding VIFs of 1.04 and 1.09 also suggest limited linear collinearity.

We next examine dependence between coefficient
estimates by calculating their correlations
across joint posterior draws. In Stage~1,
the correlation between $\beta_{\mathrm{CEI}}$
and $\beta_{D_{\mathrm{norm}}}$ is $-0.389$,
indicating a moderate trade-off: larger CEI
coefficients tend to accompany smaller demand
coefficients. In Stage~2, the correlations
are 0.106 between the event-type coefficients
($\gamma_{\mathrm{CEI}}$ and
$\gamma_{D_{\mathrm{norm}}}$) and 0.248
between the positive-capacity coefficients
($\delta_{\mathrm{CEI}}$ and
$\delta_{D_{\mathrm{norm}}}$).
These low to moderate correlations show some
dependence between estimates but no strong
linear posterior association.

Parameter stability is assessed through
leave-one-winter-out refitting and prior
sensitivity analysis. The first one evaluates
the influence of individual winters, while
the latter one examines the influence of prior
assumptions by halving or doubling slope-prior
standard deviations using the complete dataset. Changes in coefficient estimates are expected
when observations are removed. The assessment
therefore focuses on whether the direction
and uncertainty of the estimated associations
remain consistent, rather than whether the
numerical estimates remain unchanged. In Stage~1, the CEI coefficient
remains positive across these checks, with
posterior means between 1.18 and 1.31 across
winter omissions. Excluding winter~2014
reduces the demand estimate from 5.29 to
2.08, and its 95\% credible interval includes
zero. In Stage~2, excluding winter~2014 causes
the positive-capacity CEI interval to include
zero, while excluding December~2022 events
does the same for the event-type CEI interval.
The positive-capacity demand interval also
includes zero under narrower slope priors.
These findings show sensitivity to particular
winter groups and prior choices despite
limited collinearity. The estimated associations
of CEI and demand, each adjusted for the other,
should therefore be interpreted with this
sensitivity in mind based on the current dataset.

\subsection{Model Validation}

\begin{figure}[htbp]
\centering
\begin{subfigure}{0.46\textwidth}
    \centering
    \includegraphics[width=\textwidth]{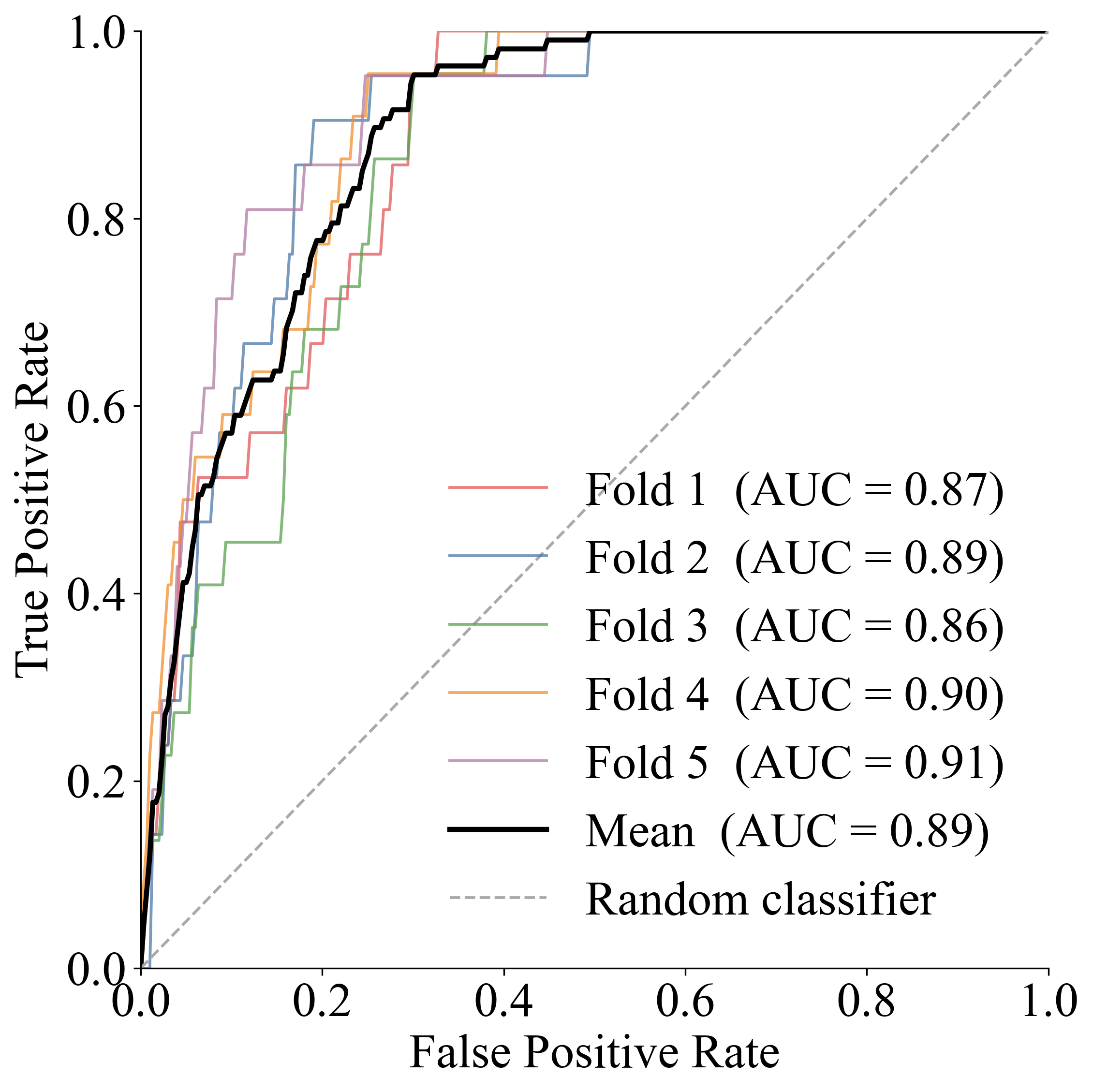}
    \caption{}
    \label{fig:s1_roc_cv}
\end{subfigure}
\hfill
\begin{subfigure}{0.50 \textwidth}
    \centering
    \includegraphics[width=\textwidth]{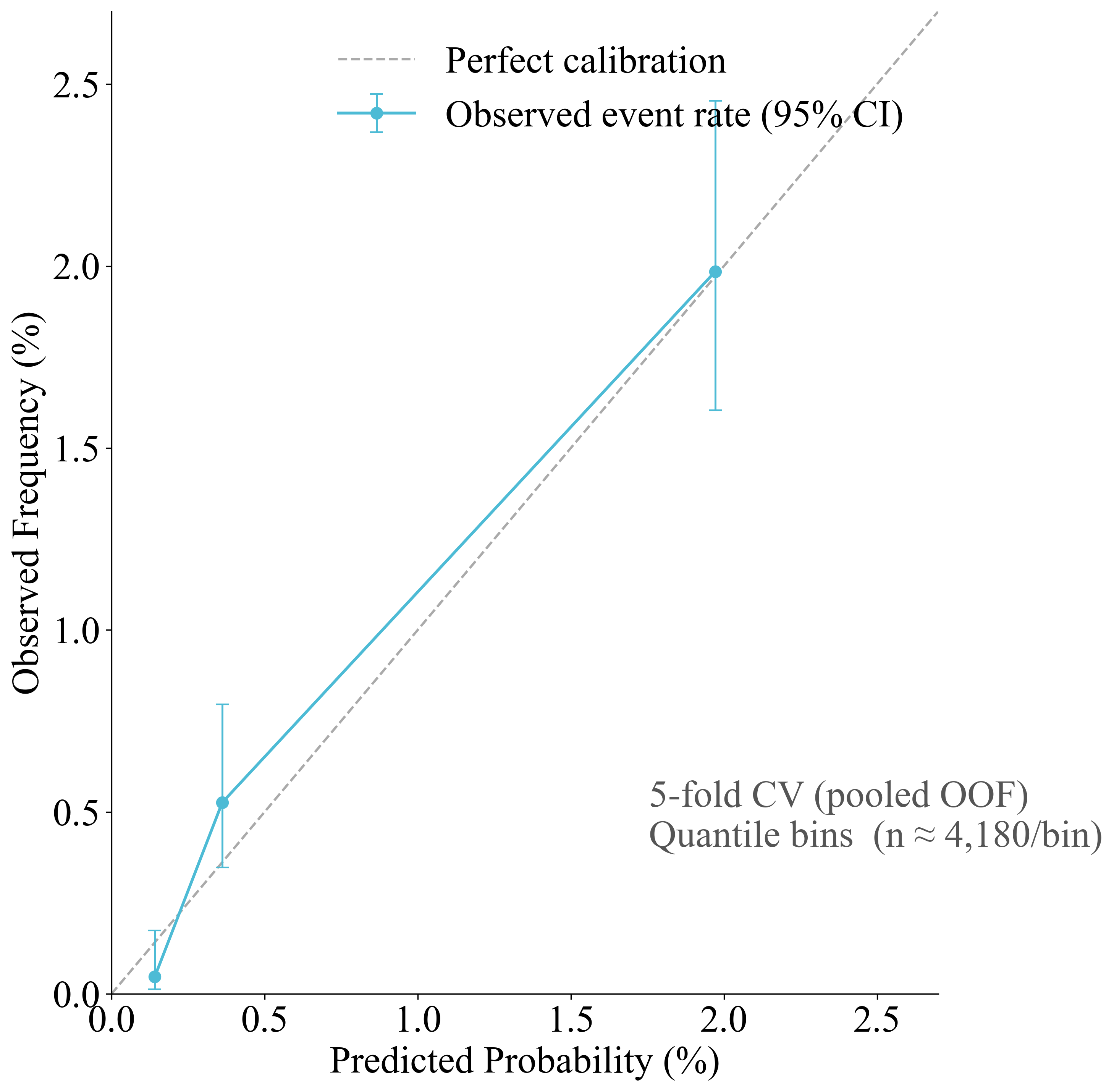}
    \caption{}
    \label{fig:s1_cal_cv}
\end{subfigure}
\caption{Stage~1 model validation: (a) ROC curves and (b) calibration
curve from five-fold stratified cross-validation.}
\label{fig:s1_validation}
\end{figure}

Stage~1 is evaluated using 5-fold stratified cross-validation, 
with stratification on the event indicator $E_t$ to preserve the 
empirical event rate in each fold. The area under the ROC curve 
ranges from 0.85 to 0.91 across folds (mean $0.89 \pm 0.02$), 
indicating consistent discrimination between event and non-event 
hours (Fig~\ref{fig:s1_roc_cv}). Calibration curve is also assessed on pooled out-of-fold predictions. As shown in Fig.~\ref{fig:s1_cal_cv} , all probability bins lie close to the perfect-calibration diagonal, confirming that predicted probabilities accurately reflect observed 
event frequencies.

\begin{figure}[htbp]
\centering
\begin{subfigure}{0.44\textwidth}
    \centering
    \includegraphics[width=\textwidth]{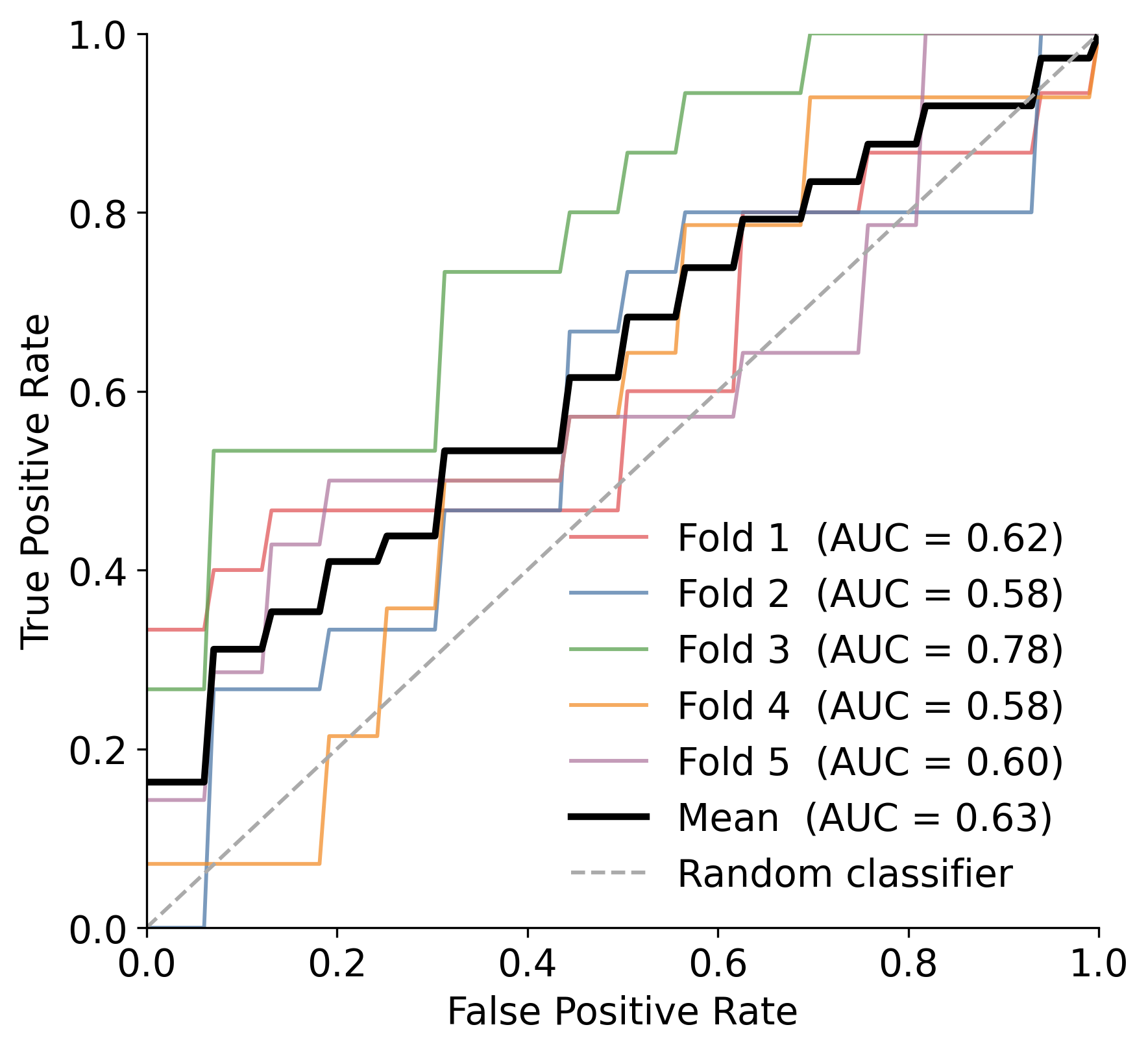}
    \caption{}
    \label{fig:s2_roc_cv}
\end{subfigure}
\hfill
\begin{subfigure}{0.52\textwidth}
    \centering
    \includegraphics[width=\textwidth]{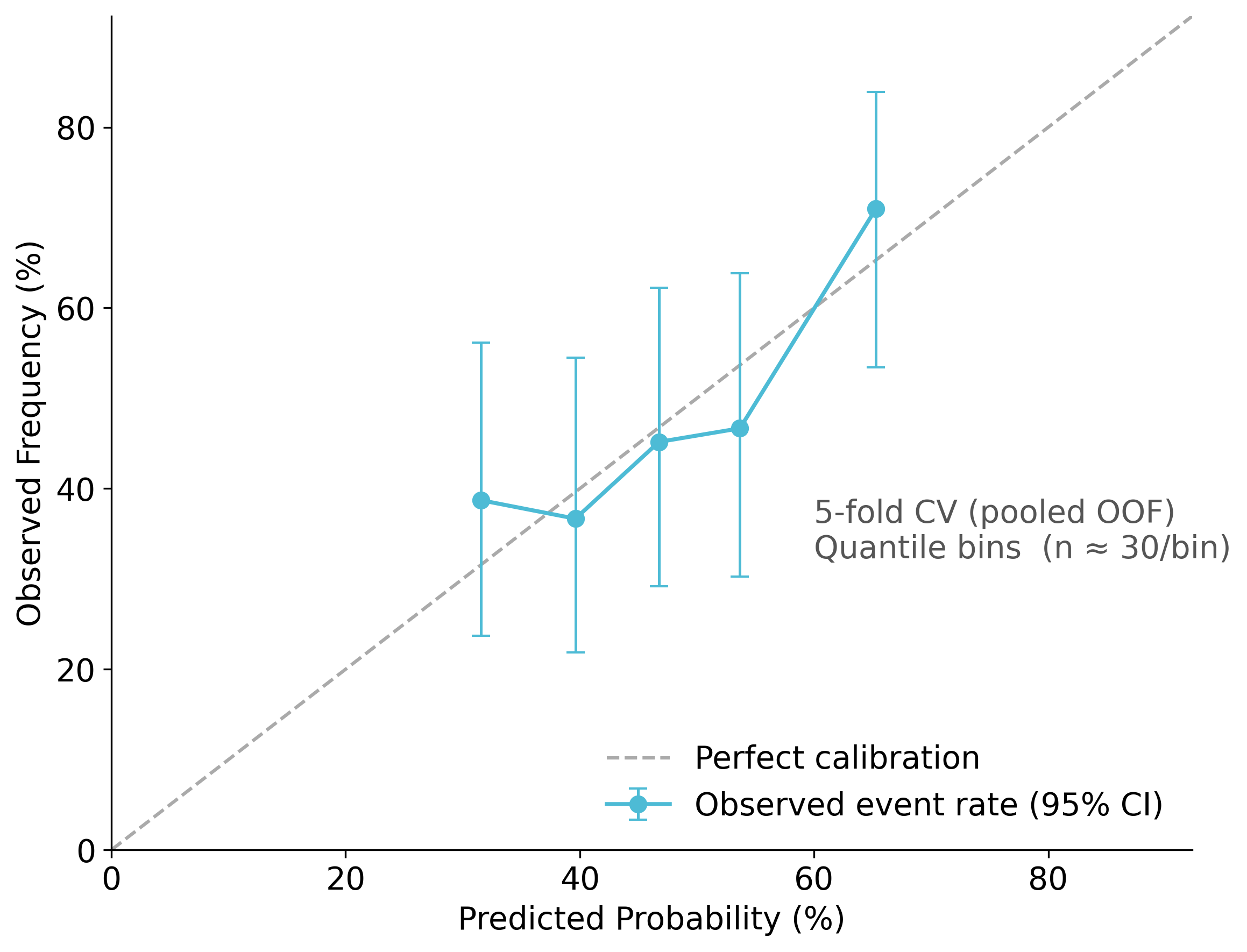}
    \caption{}
    \label{fig:s2_cal_cv}
\end{subfigure}

\vspace{0.4cm}

\begin{subfigure}{0.48\textwidth}
    \centering
    \includegraphics[width=\textwidth]{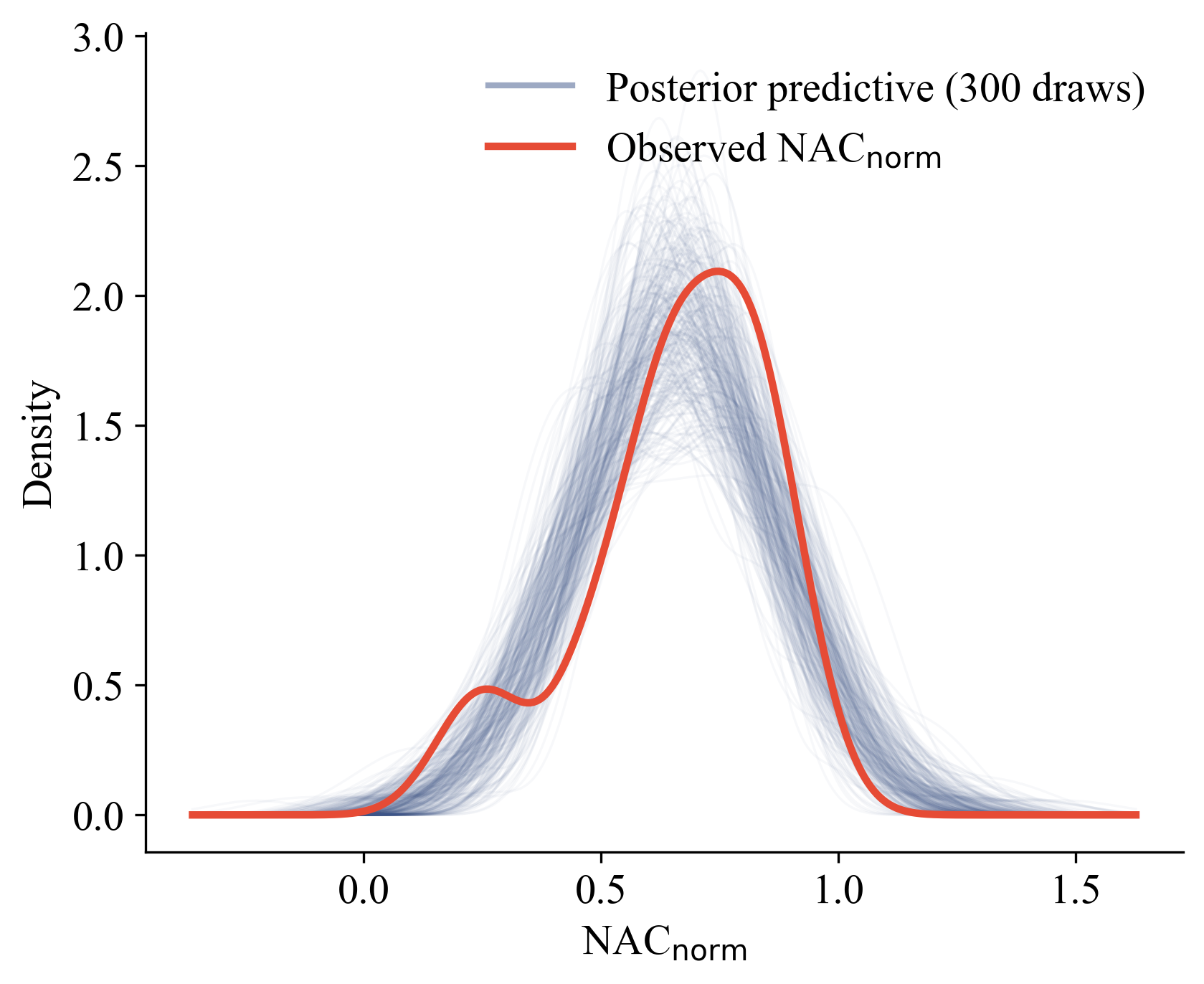}
    \caption{}
    \label{fig:s2_ppc}
\end{subfigure}
\hfill
\begin{subfigure}{0.48\textwidth}
    \centering
    \includegraphics[width=\textwidth]{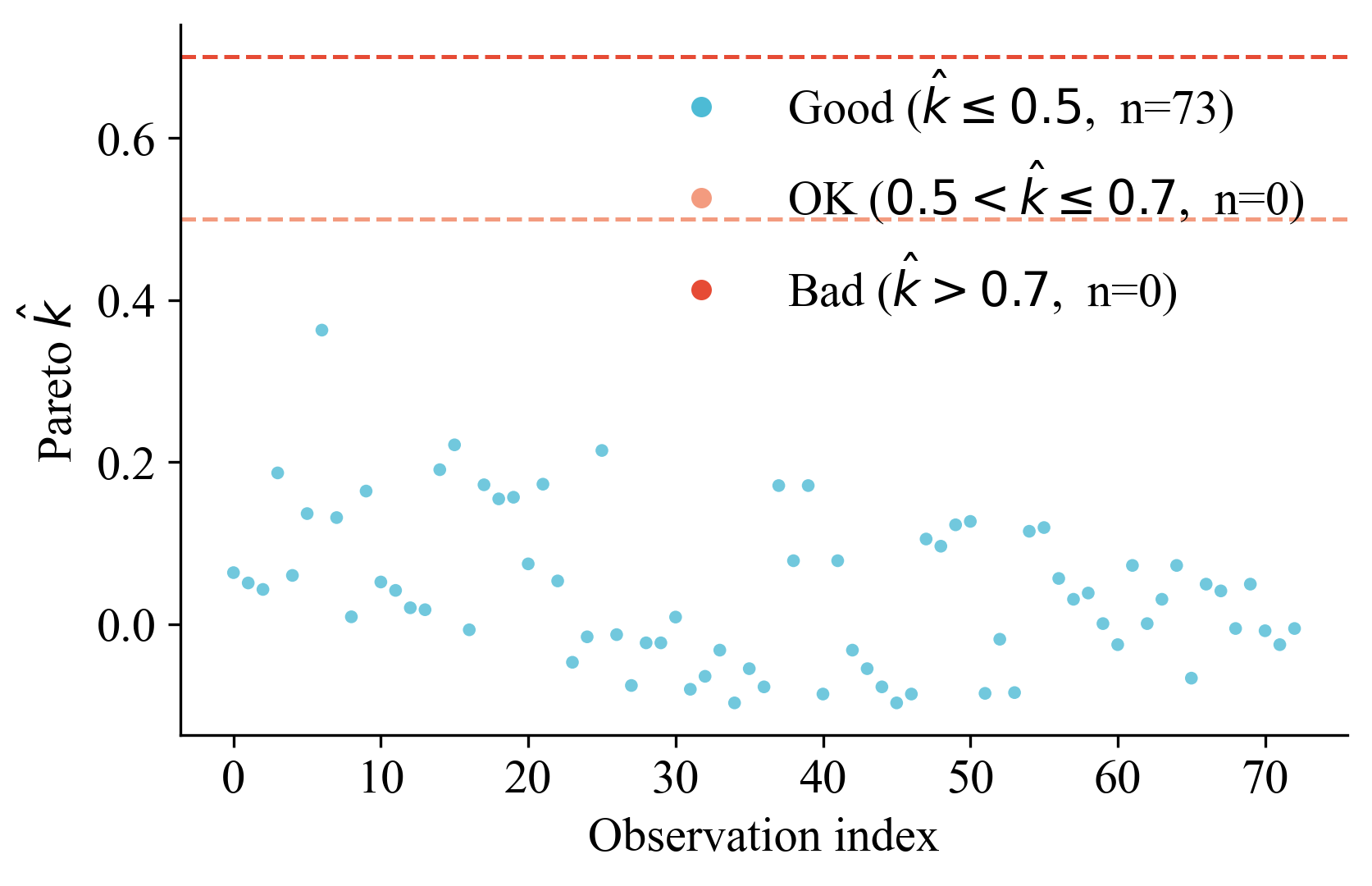}
    \caption{}
    \label{fig:s2_pareto}
\end{subfigure}

\caption{Stage~2 model validation. (a) ROC curves and (b)
calibration curve from five-fold stratified cross-validation of the
logistic sub-model. (c) Posterior predictive check and (d)
Pareto-$\hat{k}$ diagnostics for the Bayesian linear sub-model.}
\label{fig:s2_validation}
\end{figure}

For Stage~2, the Part~1 logistic sub-model is evaluated using five-fold stratified cross-validation ($K=5$) to preserve the empirical class balance (derated vs.\ full-outage) in every fold. Classification ability is assessed via the area under the ROC curve (AUC), computed separately for each fold and averaged (Fig.~\ref{fig:s2_roc_cv}). AUC values range from 0.58 to 0.81 across folds (mean $= 0.64 \pm 0.09$), reflecting moderate discrimination
between full-outage ($\mathrm{NAC} = 0$) and partial-capacity ($\mathrm{NAC} > 0$) events considering limited number of data available. Moreover, calibration is assessed on the pooled out-of-fold (OOF) predicted probabilities using five equal-frequency (quantile) bins with Wilson score 95\% confidence intervals, following the same procedure as Stage~1 (Fig.~\ref{fig:s2_cal_cv}). The small bin size reflects the limited number of confirmed contingency-event records available at this stage of the pipeline. Despite wide confidence intervals, all points track the perfect-calibration diagonal closely. 

In Part~2 linear sub-model, model adequacy is assessed using posterior predictive checks (PPC) and Pareto-smoothed importance-sampling leave-one-out cross-validation (PSIS-LOO-CV). Fig.~\ref{fig:s2_ppc} shows that the 300 posterior predictive draws closely reproduce the observed distribution of $\mathrm{NAC_{norm}}$, indicating that the model captures the principal characteristics of the data. Fig.~\ref{fig:s2_pareto} presents the corresponding Pareto-$\hat{k}$ diagnostics. All 73 observations satisfy $\hat{k}\le0.5$, placing them in the Good reliability region and confirming that the PSIS-LOO-CV approximation is reliable for all observations. Collectively, these diagnostics indicate that the model provides an adequate representation of the observed capacity degradation behavior.

The combined part of Stage~2 formulation is evaluated using five-fold out-of-fold cross-validation across all 153 contingency events. Both components use
the same training folds, with the linear component fitted exclusively on positive derates ($\mathrm{NAC}_{\mathrm{norm}} > 0$). For each held-out event, CEI and normalized demand are supplied to both components, and expected net available capacity conditional
on an event occurring is calculated via Eq.~\eqref{eq:s2_combined}. The resulting RMSE is 0.342, compared with
0.356 for the training-mean baseline,
indicating modest improvement.

\begin{figure}[htbp]
\centering
\includegraphics[width=0.55\textwidth]{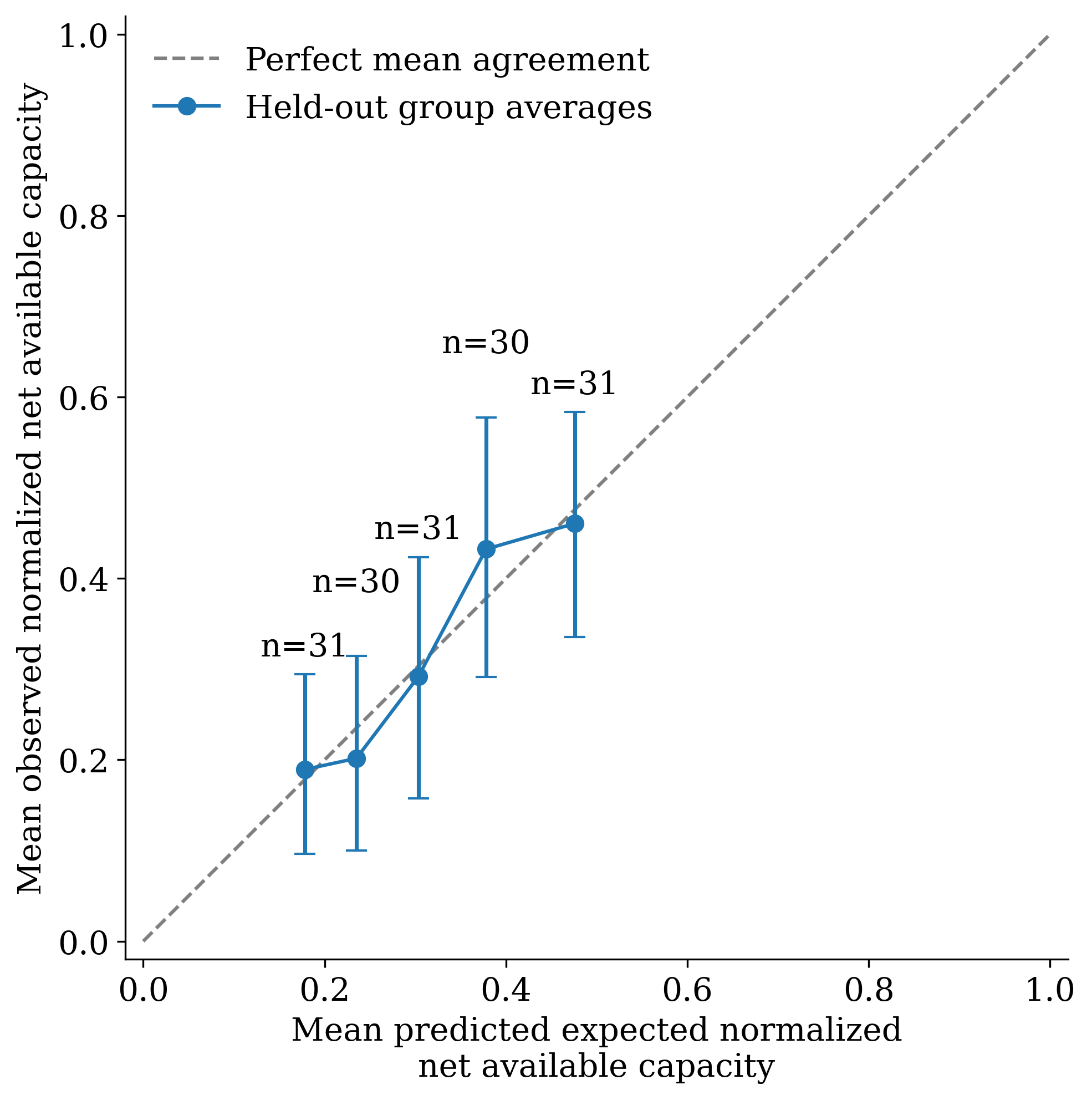}
\caption{
Out-of-fold binned calibration of combined Stage~2 capacity predictions across five equal-frequency quintiles ($n \approx 30\text{--}31$ events per bin).
}
\label{fig:s2_calibration}
\end{figure}

Figure~\ref{fig:s2_calibration} compares mean
predicted and observed capacity after grouping
the held-out records into five approximately
equal-sized bins ordered by predicted capacity
(30--31 records per bin).
The group averages lie near the 1:1 line,
and the 95\% bootstrap intervals for observed
means include the corresponding predicted means.
The agreement supports calibration of group
averages, although the wide intervals limit
the precision of this assessment.

We also examine whether held-out residuals,
defined as observed minus predicted normalized
net available capacity, vary with either input.
Pearson correlations are $-0.017$ with CEI
(95\% bootstrap interval: $[-0.171,0.138]$)
and $-0.042$ with demand
($[-0.213,0.135]$).
The corresponding Spearman correlations are
$0.109$ and $0.001$.
These results suggest little linear or
monotonic association between residuals and
either input. Together with the binned calibration results,
these findings support agreement between
predicted and observed group-average capacity.
Individual-event predictions remain less
precise, which reflects the limited available data
and the use of aggregated inputs that do not
capture plant-specific conditions.

\begin{figure}[htbp]
\centering
\begin{subfigure}{0.48\textwidth}
    \centering
    \includegraphics[width=\textwidth]{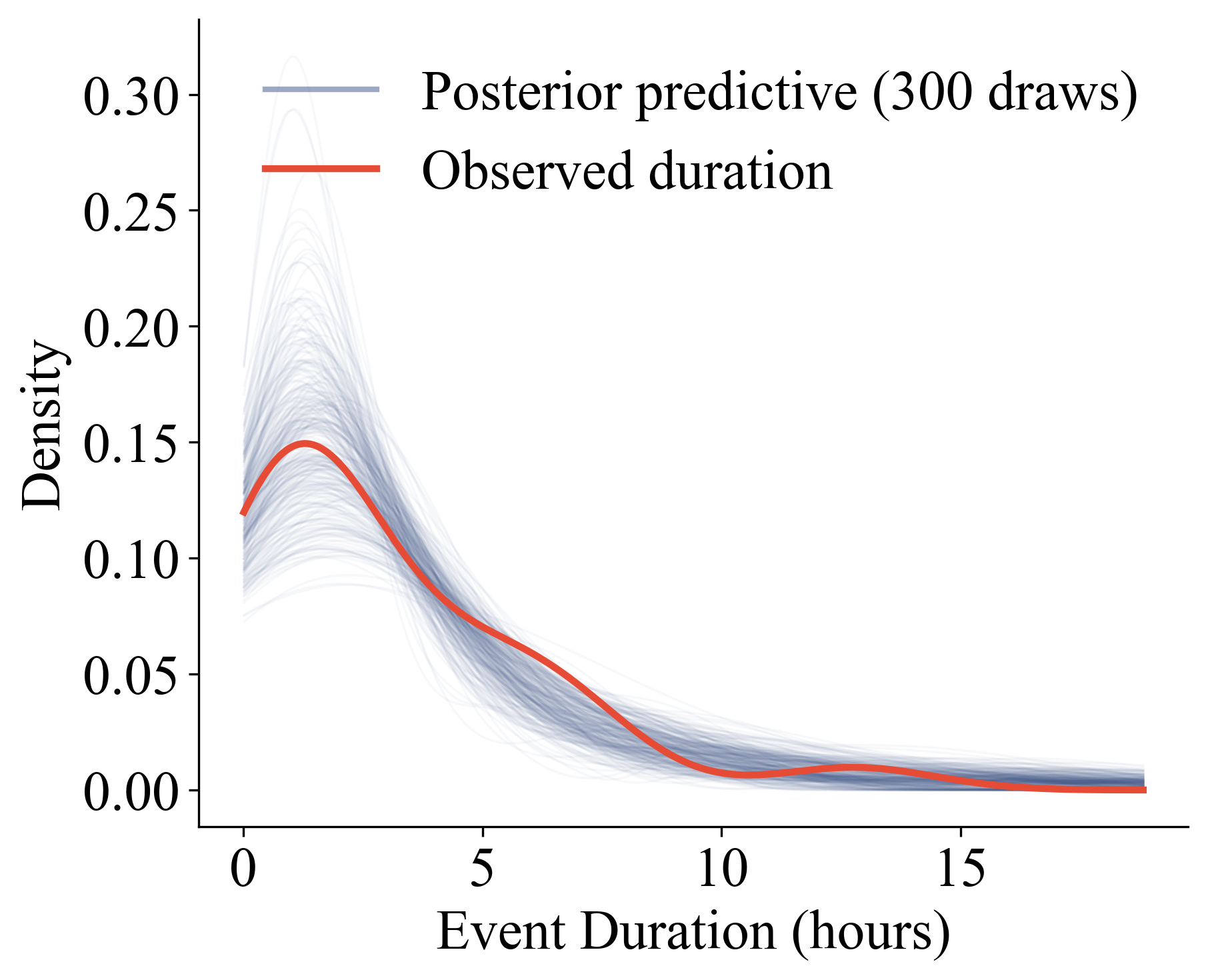}
    \caption{}
    \label{fig:s3_derate_ppc}
\end{subfigure}
\begin{subfigure}{0.48\textwidth}
    \centering
    \includegraphics[width=\textwidth]{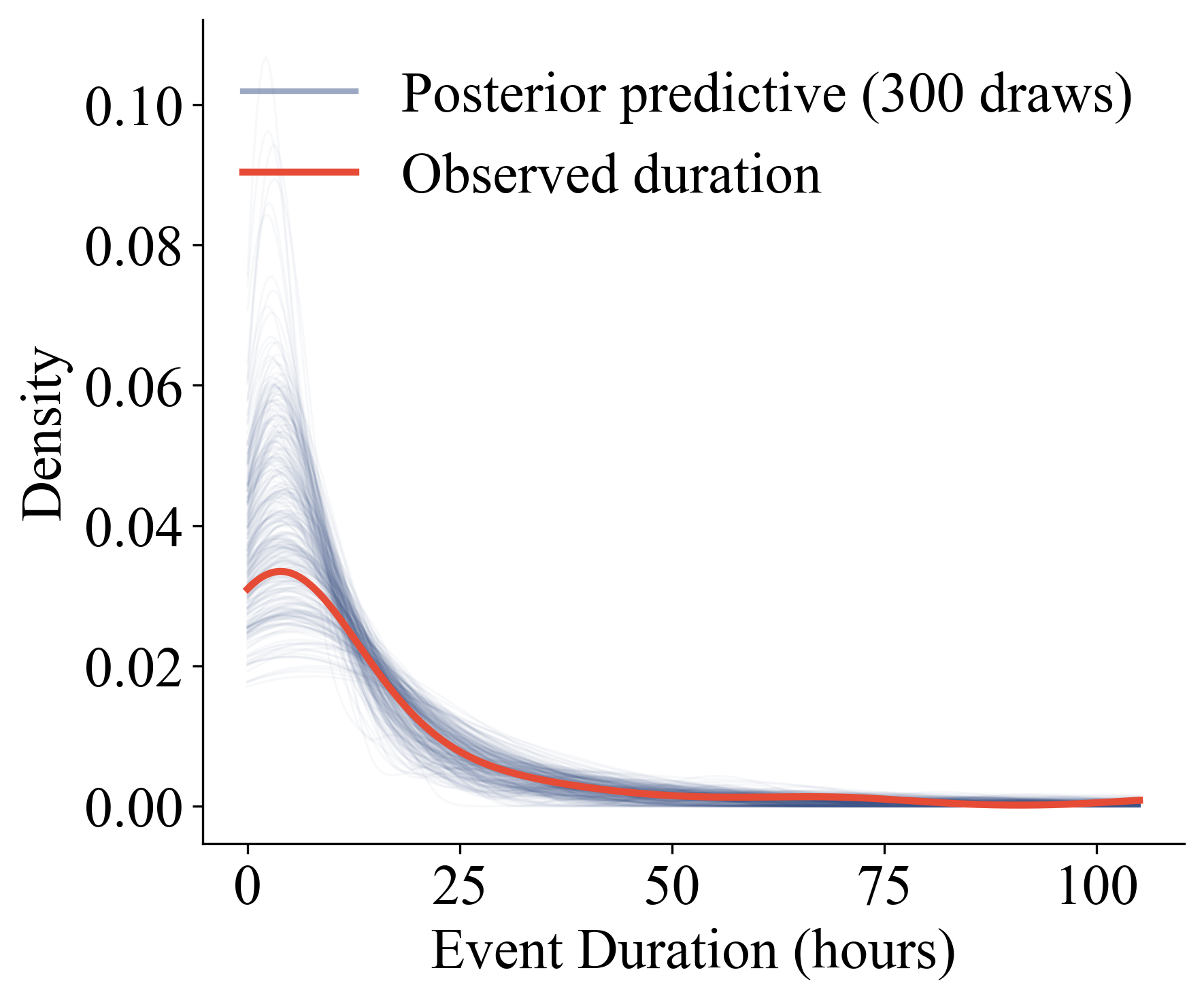}
    \caption{}
    \label{fig:s3_outage_ppc}
\end{subfigure}
\hfill
\caption{Stage~3 posterior predictive checks for (a) partial derating and (b) full outage duration models.}
\label{fig:s3_validation}
\end{figure}

Both Stage~3 sub-models are assessed using posterior predictive 
checks (PPC), consistent with the approach used for the Stage~2 
linear sub-model. For each model, 300 draws are generated from 
the combined posterior and compared to the observed duration 
distribution via kernel density estimation.  For the partial derating model (Fig.~\ref{fig:s3_derate_ppc}), the predictive draws reproduce the main body of the observed distribution (0--8 hours) well. A small secondary mode near 10 hours in the observed data is not fully reproduced by the predictive draws. For the full outage model (Fig.~\ref{fig:s3_outage_ppc}), the  observed duration distribution lies within the posterior predictive envelope across the full range, indicating that the  intercept-only lognormal specification adequately captures the  right-skewed event duration distribution of full outage events.

\begin{figure*}[t!]
    \centering
    \includegraphics[width=\textwidth]{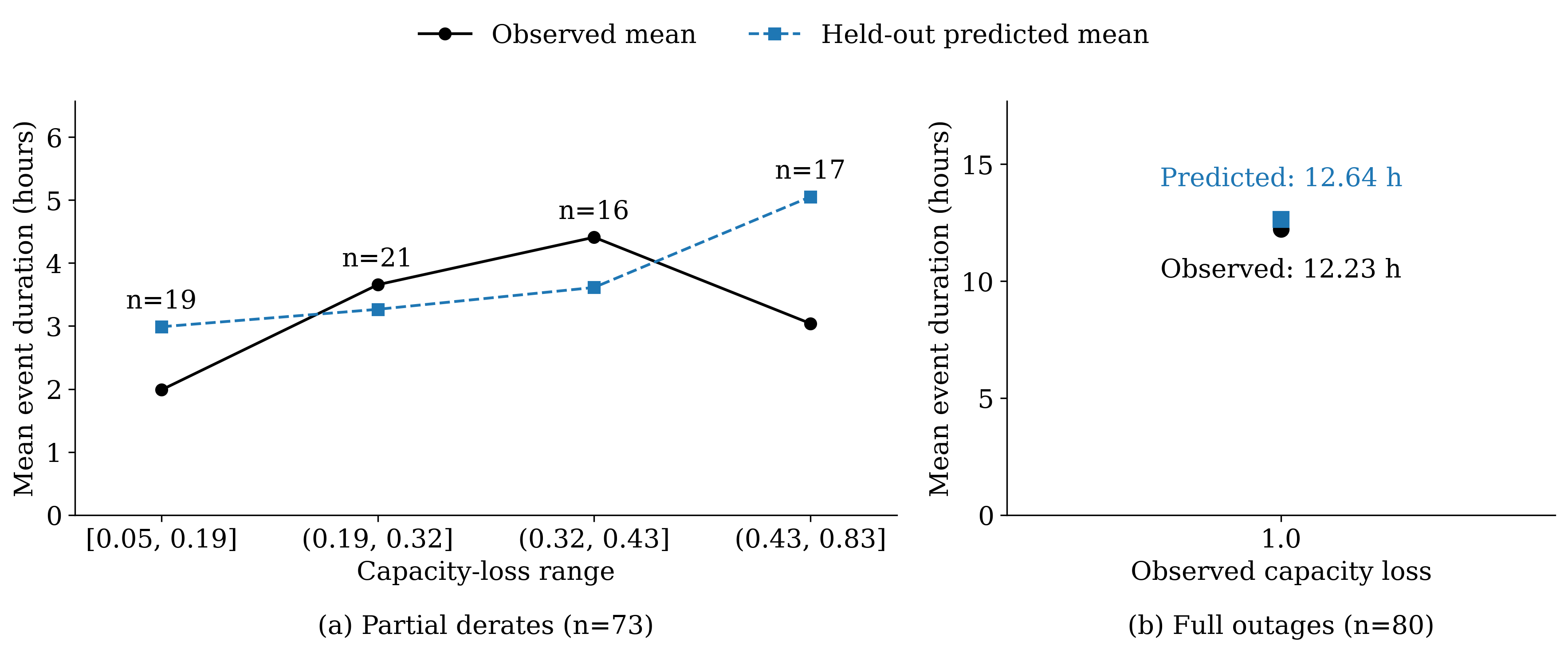}
    \caption{
    Observed and out-of-fold predicted mean
    durations for (a) partial derates grouped
    by capacity loss and (b) full outages.
    Labels indicate record counts.
    }
    \label{fig:s3_conditional}
\end{figure*}

We also evaluate Stage~3 duration predictions conditional on capacity loss using five-fold out-of-fold
predictions (Fig.~\ref{fig:s3_conditional}).
For each held-out partial derate, duration is
predicted using its observed capacity loss.
The 73 records are grouped into four capacity-loss
ranges shown in the figure, containing
19, 21, 16, and 17 records.
Predicted mean duration increases with capacity
loss, whereas observed means increase through
the first three groups before declining.
The highest-loss group is overpredicted
(5.05~h versus 3.04~h observed).
Overall RMSE is 4.09~h, close to the
training-mean baseline of 3.94~h.
The 95\% predictive intervals cover 95.9\%
of records, although their average width
of 18.17~h indicates substantial uncertainty.
For full outages, capacity loss is fixed
at unity. The predicted mean is close to
the observed mean (12.64~h versus 12.23~h),
with RMSE of 21.97~h and 95.0\% interval
coverage across 80 records.
These checks show agreement in full-outage
mean duration while identifying differences
across partial-derate severity levels and
limited accuracy for available dataset.

\section{Discussion and Conclusions}
 
This study has several limitations that shape how the results should be interpreted. First, GADS records available to us do not include plant locations or unit identifiers, so generator contingencies cannot be matched to local weather, fuel supply conditions or loading conditions. Both temperature and demand are therefore represented at the state or zonal level, which smooths out local variation and adds uncertainty to model inputs that plant-level data could help reduce it. State-level inputs combine generators with
different local weather, winterization measures,
and fuel supply conditions. The minimum
temperature across selected stations may differ
from the temperature at an affected plant,
while normalized system demand does not measure
its actual loading. These differences can affect
the estimated relationships and their application
to individual generators. The reported credible
intervals describe parameter uncertainty under
the specified models. These intervals do not separately
identify the contribution of data aggregation.
Quantifying that contribution would require
linked plant-level observations for comparison
with the aggregated inputs and predictions.

Second, cause codes in GADS are assigned by plant operators and vary widely in specificity, so identifying winter-driven events requires conservative filtering that may exclude events whose cause codes do not explicitly reference weather-related conditions. The resulting sample data are limited across all three modeling stages, and the true number of cold-driven contingencies is likely higher than what the current cause-code structure allows us to recover. The small sample limits the precision of the estimations.

Third, cold exposure index and normalized demand do not capture the full range of winter failure pathways. For instance, gas supply interruption, differences in plant winterization practice all contribute to observed outcomes but are absent from available records. In addition, event duration can be influenced by factors such as damage type, repair logistics, and resource availability that are not available publicly. For decisions concerning individual generators,
plant-specific assessments informed by equipment
condition and failure mechanisms would be more
appropriate when the necessary records are
available. The current framework provides a
regional statistical baseline that could
complement those assessments.  However, its suitability
for operational use requires further validation.

Fourth, validation in New York does not establish
predictive performance in other regions.
Differences in climate, generation technology,
fuel supply arrangements, and winterization
practices may change the estimated relationships.
Although the three-stage structure can be
applied elsewhere, the New York coefficients
should not be assumed transferable without
regional calibration and validation. Comparisons
across states would help determine which
relationships are consistent and which require
region-specific estimation.

Fifth, predictive performance in this study is conducted stage by stage. However, an end-to-end evaluation of the full framework, in which event initiation, capacity, and duration are
simulated together across a held-out winter and compared against observed
system-level capacity loss, is not performed here. Such an evaluation would
score a predictive distribution against a single realized value for each of
the ten available winters. The leave-one-winter-out analysis reported in
Section~4.4 shows that individual coefficients are sensitive to the inclusion of
specific winters. A score computed over the same ten winters would therefore
reflect this sensitivity rather than the predictive structure of the combined
framework. End-to-end validation of the combined framework with richer data remains an important direction for future work.

Despite these constraints, the three modeling stages together reveal a consistent and physically interpretable picture of winter generator risk, recovering systematic relationships between cold exposure, system demand, event probability, capacity degradation, and event duration from aggregated data alone. The principal contribution of this study is not predictive precision but the modular framework. Each stage addresses a distinct question, providing a richer representation of winter reliability risk and a direct interface with resource adequacy tools. To our knowledge, this is the first study to build a winter-storm-focused, multi-stage probabilistic framework for gas-fired generator failure, performance degradation, and failure event duration using available contingency event data.

The three-stage structure is also directly compatible with hourly Monte Carlo simulation for resource adequacy assessment. At each simulation hour, Stage~1 determines whether a contingency event is triggered given the prevailing ${\mathrm{CEI}}$ and $D_{\mathrm{norm}}$. If triggered, Stage~2 provides the expected capacity retained and Stage~3 draws a event duration. The generator then remains in the degraded state for that duration before being re-evaluated, preserving temporal consistency across consecutive hours without re-sampling during an ongoing event. This sequential logic allows the framework to be embedded directly into existing power system simulation tools to translate weather-driven generator uncertainty into system-level reliability metrics during major winter events.

While the current framework is developed on regionally aggregated records from New York, its modular three-stage conditioning architecture is extensible to granular plant-level failure mechanisms as operational data become accessible. In winter conditions, gas-fired generator vulnerabilities predominantly arise from two distinct physical pathways: mechanical freeze-ups (such as frozen instrumentation lines, valves, and water intake systems) and fuel delivery disruptions (such as pipeline pressure deficits, wellhead freeze-offs, or curtailments of non-firm transportation contracts). Incorporating these pathways requires localized operational and environmental data, including plant-level microclimate observations, winterization status, pipeline operational flow orders, on-site dual-fuel switching capability, etc. These mechanisms can be integrated across the framework through targeted enhancements without disrupting the sequential modeling logic. In Stage~1, the binary initiation model can be expanded to a multinomial logit formulation to differentiate between mechanical trips, fuel-supply curtailments, and cold-start ignition failures following offline periods. In Stage~2, differences in unit age, size, and turbine type can be captured through a hierarchical model, allowing individual plants to have distinct weather sensitivities while sharing information across the broader fleet. In Stage~3, event duration can be grouped by failure type, separating onsite quick mechanical fixes from longer outages driven by natural gas supply disruptions or vice versa. Furthermore, logistics constraints, such as repair crew availability, hazardous site road access, and regional outage concurrency, can enter Stage~3 as duration covariates. In this manner, the framework scales naturally from regional aggregate planning to plant-level operational reliability modeling. While plant-specific models are better suited for unit-level decisions, the posterior distributions estimated here can serve as informative priors for such models.

This study shows how available contingency,
weather, and demand records can be combined
to estimate winter-driven event probability,
expected net available capacity, and event
duration within a three-stage Bayesian framework.
The New York application provides a starting
point for examining how cold exposure and
system demand relate to generator contingencies
when plant-level records are unavailable. Although aggregated data allow these relationships
to be examined, application to
individual generators and in other regions
requires further calibration and validation. This highlights the
urgent need for higher-fidelity data sharing. With access to plant-level weather, operating,
and failure records, the framework could support
more representative assessments of generator
vulnerability and inform winter preparedness
and reliability planning.

\appendix
\section{Posterior parameter estimates (all stages)}
\label{app1}

\begin{table}[htbp]
\centering
\caption{Posterior parameter estimates for the three-stage Bayesian
framework}
\label{tab:posteriors}
\begin{tabular}{llcc}
\toprule
\textbf{Stage} & \textbf{Parameter} & \textbf{Mean} & \textbf{95\% CI} \\
\midrule
\multirow{3}{*}{\shortstack[l]{Stage~1\\Logistic}}
  & $\beta_0$           & $-9.484$ & $[-11.318,\ -7.766]$ \\
  & $\beta_{\mathrm{CEI}}$           & $1.173$  & $[0.929,\ 1.392]$    \\
  & $\beta_{D_{\mathrm{norm}}}$      & $5.299$  & $[2.443,\ 8.087]$    \\
\midrule
\multirow{2}{*}{\shortstack[l]{Stage~2\\Logistic}}
  & $\gamma_{\mathrm{CEI}}$          & $-0.577$ & $[-0.998,\ -0.175]$  \\
  & $\gamma_{D_{\mathrm{norm}}}$     & $-0.773$ & $[-2.314,\ 0.691]$   \\
\midrule
\multirow{2}{*}{\shortstack[l]{Stage~2\\Linear}}
  & $\delta_{\mathrm{CEI}}$          & $-0.061$ & $[-0.108,\ -0.014]$  \\
  & $\delta_{D_{\mathrm{norm}}}$     & $-0.107$ & $[-0.256,\ 0.047]$   \\
\midrule
\multirow{2}{*}{\shortstack[l]{Stage~3\\Full outage}}
  & $\theta_0^{\mathrm{out}}$        & $1.495$  & $[1.114,\ 1.882]$    \\
  & $\sigma_{\mathrm{out}}$          & $1.411$  & $[1.174,\ 1.702]$    \\
\midrule
\multirow{3}{*}{\shortstack[l]{Stage~3\\Partial derate}}
  & $\theta_0^{\mathrm{der}}$        & $0.200$  & $[-0.413,\ 0.797]$   \\
  & $\theta_1^{\mathrm{der}}$        & $1.161$  & $[-0.366,\ 2.715]$   \\
  & $\sigma_{\mathrm{der}}$          & $1.139$  & $[0.947,\ 1.387]$    \\
\bottomrule
\end{tabular}
\end{table}







\section*{Data availability}
The Generating Availability Data System (GADS) dataset from the North American Electric Reliability Corporation (NERC) is not directly available online but can be accessed upon request to NERC~\cite{GADS}. Redistribution of the data used in this study requires permission from NERC. Weather data from NOAA ISD, generator location data from eGRID, and demand data from NYISO are available~\cite{NOAA_ISD, egrid, nyiso}.

\section*{Code availability}
All Python code used to implement the three-stage Bayesian framework is available at \url{https://github.com/sajjad30148/winter_gas_risk_model}. Full reproduction of the results requires access to the GADS dataset.

\section*{Acknowledgment}
The authors would like to thank Dr. Chiara Lo Prete ( Pennsylvania State University), Dr. Burçin Ünel (New York University) and Dr. Ashish Radhakrishnan (Pennsylvania State University) for their support and valuable feedback; Mark Lauby and Svetlana Ekisheva from North American Electric Reliability Corporation (NERC) for providing access to the GADS data used in this study.

\section*{Funding}
This work was supported by National Science Foundation (NSF) SAI Grant \#2324544.

\section*{CRediT authorship contribution statement}
\textbf{Sajjad Uddin Mahmud:} Conceptualization, Data curation, Methodology, 
Software, Formal analysis, Validation, Visualization, Writing -- original draft, Writing -- review \& editing.

\textbf{Anamika Dubey:} Conceptualization, Methodology, Supervision, Project 
administration, Resources, Funding acquisition, Writing -- review \& editing.

\section*{Declaration of Competing Interest}
The authors declare that they have no known competing financial interests or personal relationships that could have appeared to influence the work reported in this paper.
    
\bibliographystyle{elsarticle-num}
\bibliography{references}

\end{document}